\documentclass[runningheads]{llncs}

\usepackage{eccv}

\usepackage{eccvabbrv}

\usepackage{graphicx}
\usepackage{booktabs}
\usepackage{amssymb}
\usepackage{amsmath}
\usepackage{comment}
\usepackage{array}
\usepackage{multirow}
\usepackage{tabularx}
\usepackage{xcolor}
\usepackage{diagbox}

\newcolumntype{Y}{>{\centering\arraybackslash}X}
\newcommand{\key}[1]{[\makebox[1.0em][c]{#1}]}
\usepackage[accsupp]{axessibility}  

\usepackage{hyperref}

\usepackage{orcidlink}

\newcommand{\tick}{\raisebox{0.5ex}{\scalebox{0.7}{$\sqrt{}$}}}
\newcommand{\ntick}{\raisebox{-0.1ex}{$\times$}}
\newcommand{\halftick}{\raisebox{0.5ex}{\scalebox{0.7}{$\sqrt{}\mkern-9mu{\smallsetminus}$}}}

\newcommand{\methodName}{UNICA}

\begin{document}

\title{UNICA: A Unified Neural Framework for Controllable 3D Avatars\vspace{-0.7em}} 

\author{Jiahe Zhu\and
        Xinyao Wang\and
        Yiyu Zhuang\and
        Yanwen Wang\and
        Jing Tian\and
        Yao Yao\and
        Hao Zhu\thanks{Corresponding author}\vspace{-0.1in}}

\authorrunning{J. Zhu \etal}

\institute{Nanjing University, Nanjing, China\\
State Key Laboratory of Novel Software Technology, China
}

\maketitle
\vspace{-0.2in}

\begin{figure*}[h]
	\centering
    \vspace{-0.1in}
	\includegraphics[width=\textwidth]{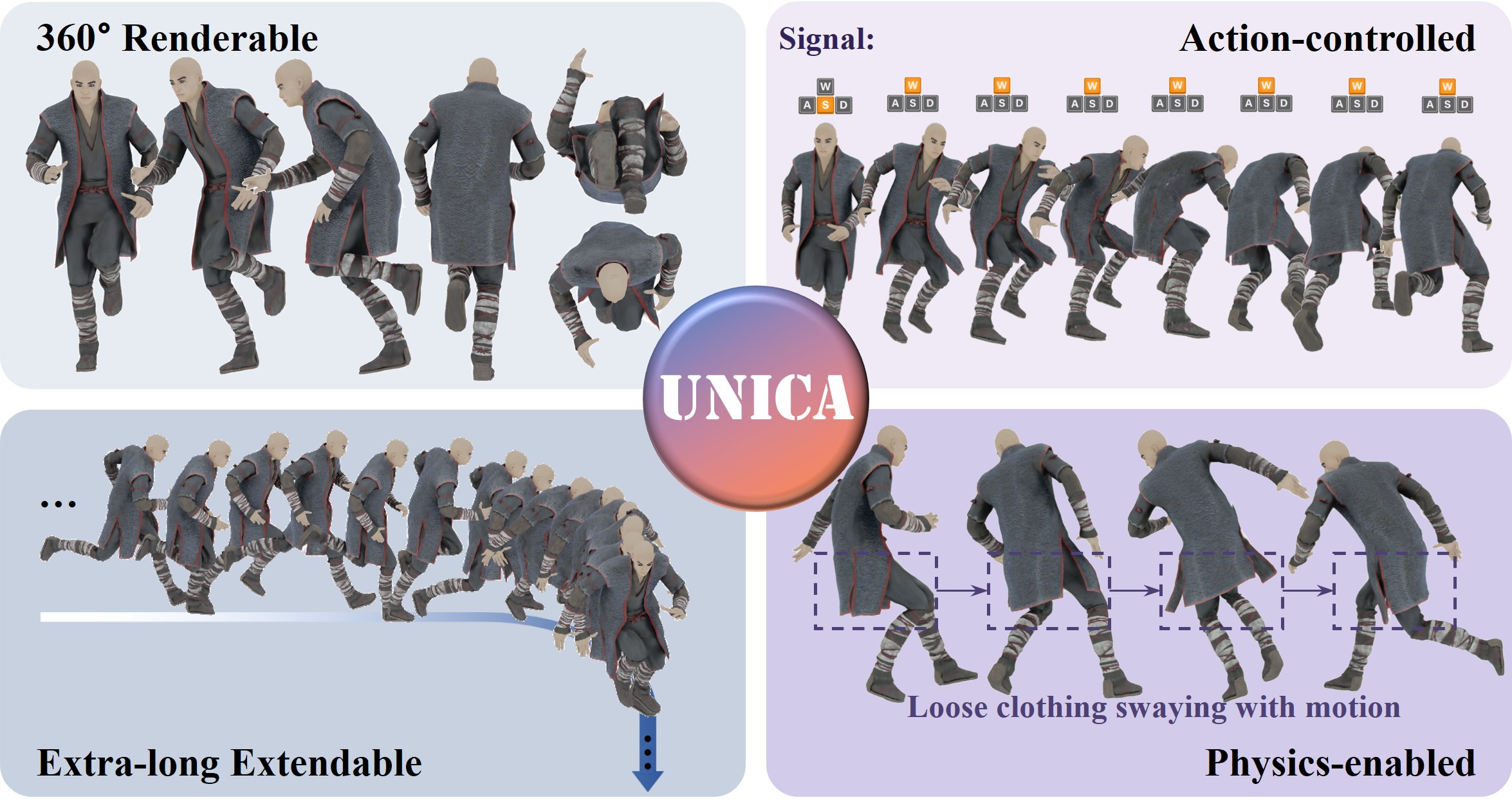}
    \vspace{-0.15in}
    \caption{
    \methodName{} is a unified model that generates action-controlled, $360^\circ$-renderable 3D avatars with dynamics. 
    For the first time, \methodName{} unifies a workflow of "\textit{motion planning, rigging, physical simulation, and rendering}" within a \textbf{single} model.}
    \vspace{-0.45in}
	\label{fig:teaser}
\end{figure*}

\begin{abstract}
Controllable 3D human avatars have found widespread applications in 3D games, the metaverse, and AR/VR scenarios. The conventional approach to creating such a 3D avatar requires a lengthy, intricate pipeline encompassing appearance modeling, motion planning, rigging, and physical simulation.
In this paper, we introduce \methodName{} (UNIfied neural Controllable Avatar), a skeleton-free generative model that unifies all avatar control components into a single neural framework. Given keyboard inputs akin to video game controls, \methodName{} generates the next frame of a 3D avatar's geometry through an action-conditioned diffusion model operating on 2D position maps. A point transformer then maps the resulting geometry to 3D Gaussian Splatting for high-fidelity free-view rendering. Our approach naturally captures hair and loose clothing dynamics without manually designed physical simulation, and supports extra-long autoregressive generation. To the best of our knowledge, \methodName{} is the first model to unify the workflow of "\textit{motion planning, rigging, physical simulation, and rendering}". Code is released at \url{https://github.com/zjh21/UNICA}.
\vspace{-0.1in}
\end{abstract}

\section{Introduction}
\label{sec:intro}
3D avatars are essential to the metaverse, video games, and film production. A defining characteristic of 3D avatars is their controllability: unlike static objects, avatars in games, telepresence, and virtual worlds must respond to instructions and execute diverse actions. Currently, the most direct and intuitive form of control is through key presses—for example, players pressing the `WASD' keys to direct an avatar's movement. At runtime, this control follows a well-established process built upon a skeletal system. First, a motion planning module determines the appropriate skeletal pose changes in response to the key input. The avatar's surface geometry is then deformed accordingly via skinning weights, which are calibrated during the rigging stage. Subsequently, physical simulation is applied to govern the dynamics of non-rigid components such as hair and loose clothing. This pipeline has served as the conventional workflow for gaming and film engines for decades. Nevertheless, its dependence on extensive manual effort from skilled artists at the authoring stage renders it prohibitively expensive.

Over the past decade, learning-based methods have automated individual stages of the avatar control pipeline, including skeleton generation~\cite{pred_animation_skeletons, unirig}, motion prediction~\cite{pfnn, deepphase}, and mesh deformation~\cite{neural_blend_shape, skinningnet}. However, most of these methods address one or two stages, leaving the remaining components to be manually designed. Human parametric models~\cite{smpl, smplh, smplx} further simplify avatar control by providing a universal skeleton with predefined skinning weights, and have been widely adopted as the foundation for reconstructing animatable avatars from multi-view videos~\cite{animatable_gaussians, taoavatar}. However, the resulting avatars are animatable but not control-ready: users cannot simply press a key and see the avatar respond; instead, animation still requires obtaining pose sequences from external sources. 

In this paper, we propose UNIfied neural Controllable Avatar (\methodName{}), a novel strategy to unify the series of tasks from ``\textit{key presses - motion planning - rigging - physical simulation - rendering}'' to a direct progression ``\textit{key presses - rendering}''.  We demonstrate that this task of driving an avatar with actions can be accomplished end-to-end solely with a single generative framework, thereby fundamentally reducing the construction costs of a controllable avatar. To train \methodName{}, we use mesh sequences of rigged avatars exported from Unreal Engine, which encompass locomotion in four directions (corresponding to the 'WASD' keys) and directional transitions. The core of \methodName{} is an action-conditioned diffusion model~\cite{ddpm} designed for autoregressive inference.  Specifically, conditioned on historical geometric context and a specific user action (e.g., `WASD' or idle), the model predicts the subsequent 3D geometry. To leverage established 2D diffusion architectures~\cite{ldm, svd}, we encode 3D geometry into 2D position maps, where each foreground pixel corresponds to a 3D coordinate. Crucially, by densely sampling the clothing surface, these position maps preserve high-frequency geometric details, enabling the model to implicitly learn and reproduce complex physical dynamics, such as cloth folding and deformation. Finally, a feed-forward Point Transformer~\cite{ptv3} maps the generated point clouds to 3D Gaussian Splatting (3DGS)~\cite{3dgs} primitives to model the avatar's comprehensive appearance.

\begin{table}[tb]
\centering
\caption{Comparison between \methodName{} and prior approaches.}
\vspace{-0.1in}
\begin{tabularx}{\textwidth}{@{\hspace{0.5em}\extracolsep{\fill}}lYYYY@{}}
\toprule
Method & $360^\circ$ Render & Key-Control & Arbitrary Len. & Physics \\
\midrule
Animatable 3D Avatar & \tick & \ntick & \tick & \halftick \\
Motion Generation & \tick & \halftick & \halftick & \ntick \\
World Model & \ntick & \tick & \tick & \tick \\
4D Generation & \tick & \ntick & \halftick & \tick \\
\methodName{} (Ours) & \tick & \tick & \tick & \tick \\
\bottomrule
\multicolumn{5}{l}{\tick~$=$ supported; \ntick~$=$ NOT supported; \halftick~$=$ not fully supported/might be extended.}
\end{tabularx}
\vspace{-0.2in}
\label{tab:comparison_with_existing}
\end{table}

As a result of this design, every output frame generated by \methodName{} constitutes a $360^\circ$-renderable 3D avatar exhibiting realistic physical dynamics. Furthermore, we introduce a progressive 4D inference scheme that enables avatar navigation in unbounded world coordinates. Experimental results demonstrate that \methodName{} supports long-term generation, synthesizing more than 1,000 frames without degrading quality. The salient features of \methodName{} are illustrated in \cref{fig:teaser}. To the best of our knowledge, this work is the first to explore end-to-end action-driven 3D neural avatars. \cref{tab:comparison_with_existing} provides a comparative analysis between \methodName{} and existing approaches addressing related sub-tasks. Leveraging these capabilities, \methodName{} facilitates the automatic creation of controllable 3D avatars from light-stage captures.

In summary, our contributions are as follows:

\begin{itemize}
    \item We introduce a new paradigm for 3D avatar control: action-driven neural avatars that respond to key presses (akin to interactive gaming characters), thereby obviating the need for motion planning, model rigging, and physical simulation.
    
    \item We propose \methodName{}, a skeleton-free framework that combines an action-conditioned diffusion model for 4D generation with a point transformer for appearance synthesis. It enables free-view rendering with natural hair and clothing dynamics, while ensuring stability during extra-long autoregressive generation.
    
    \item A specialized training strategy is developed to learn controllable dynamics directly from 4D human data. By organizing data into overlapping temporal windows and introducing progressive 4D inference, we facilitate unbounded avatar navigation over arbitrary distances.
\end{itemize}
\section{Related Work}
\label{sec:related}

\subsubsection{Animatable 3D Avatars.}
The traditional approach to creating animatable 3D avatars involves modeling geometry and rigging with a manually defined skeleton, though recent work has explored automating this process~\cite{pred_animation_skeletons, riggnet, unirig, riganything}. A rapidly advancing alternative constructs animatable avatars from multi-view videos captured in light stages~\cite{animatable_gaussians, taoavatar, gaussian_avatar}. Early works explored avatar modeling based on meshes~\cite{mesh_based_avatar_2, mesh_based_avatar_3} and Neural Radiance Field (NeRF)~\cite{smpl_based1, nerf_avatar_1, nerf_avatar_2}. Recent advances have shifted towards 3D Gaussian Splatting (3DGS)~\cite{3dgs}, an explicit alternative where Gaussian ellipsoids can be transformed with avatar motion. Animatable Gaussians~\cite{animatable_gaussians} builds a parametric template covering hair and loose clothing, driving it via LBS and rendering to position maps for Gaussian attribute prediction. TaoAvatar~\cite{taoavatar} further extends this approach by distilling the StyleUNet~\cite{style_unet} into a lightweight MLP, enabling mobile deployment. Despite these advances, such avatars remain pose-driven: animation still requires pose sequences from external sources, leaving a gap between the reconstructed representation and direct user control.

\subsubsection{Motion Generation.}
In recent years, neural network-based motion generation has emerged as an automatic way to design pose changes of avatars. Early methods predicted future poses~\cite{pose_pred_1, pose_pred_2} or interpolated between keyframes~\cite{keyframe_intepolation_1, keyframe_intepolation_2}. Learning from motion capture data~\cite{humanml3d, amass, kit_ml}, recent advances have centered on text-prompted motion generation, with the diffusion paradigm widely adopted for this task~\cite{mdm, emdm, autokeyframe}. However, the outputs of such models are fixed-length pose sequences that cannot be easily extended. A line of work more closely related to ours is neural-based online character control~\cite{pfnn, mode_adaptive_nn, neural_state_machine, local_motion_phases, deepphase}. The foundational PFNN~\cite{pfnn} draws an analogy between the intermediate states within a gait cycle and the phase of a periodic function, predicting locomotion with phase-functioned neural networks. Subsequent work extends this framework to actions beyond locomotion and transitions between different types of motion~\cite{neural_state_machine, neural_motion_graph}. However, these methods operate at the skeletal level and must be combined with separate rigging and physical simulation.

\subsubsection{World Models.}
World models are systems designed to understand and predict the evolution of an environment given historical observations~\cite{world_model_review}. Game simulation is an ongoing research direction for world models, predicting future scenes conditioned on action inputs~\cite{genie, matrix_game, matrix_game20, wham}. Closely related to our idea, GameNGen~\cite{gamengen} fits an entire first-person video game with a multi-frame diffusion model conditioned on historical frames and player actions. State-of-the-art world models~\cite{genie2, genie3, lingbot_world, yume15, hunyuan_world} have demonstrated the ability to synthesize videos across diverse domains, with some approaches including third-person human motion. However, these approaches are built upon 2D video generation models, which lack explicit 3D geometric understanding. This leads to issues such as abnormal avatar motion, appearance shift, and error accumulation.

\subsubsection{4D Generation.}
4D generation~\cite{sv4d, sv4d20, l4gm} aims to produce the temporally varying appearance of a 3D object. Early attempts~\cite{4dfy, sds4d_1, sds4d_2} adopted Score Distillation Sampling (SDS)~\cite{sds}, obtaining supervision from image~\cite{ldm, sd30, imagen}, video~\cite{svd, videofactory}, and 3D generation models~\cite{sv3d, mvdream}. However, they inherit persistent issues such as long inference times and poor spatiotemporal consistency. 
Subsequently, multi-view video generation~\cite{sv4d, sv4d20, diffusion2} has emerged as an alternative, but it requires a separate 4D reconstruction step~\cite{dnerf, 4dgs} to convert the generated videos into usable 4D assets. More recently, ShapeGen4D~\cite{shapegen4d} proposes an end-to-end 4D generative pipeline in which multiple frames of 3D latents are jointly generated and then decoded into 3D shapes. Despite these advances, action-controlled animation and streamlined inference remain unexplored.
\section{Method}
\label{sec:method}


\begin{figure*}[tb]
	\centering
	\includegraphics[width=\textwidth]{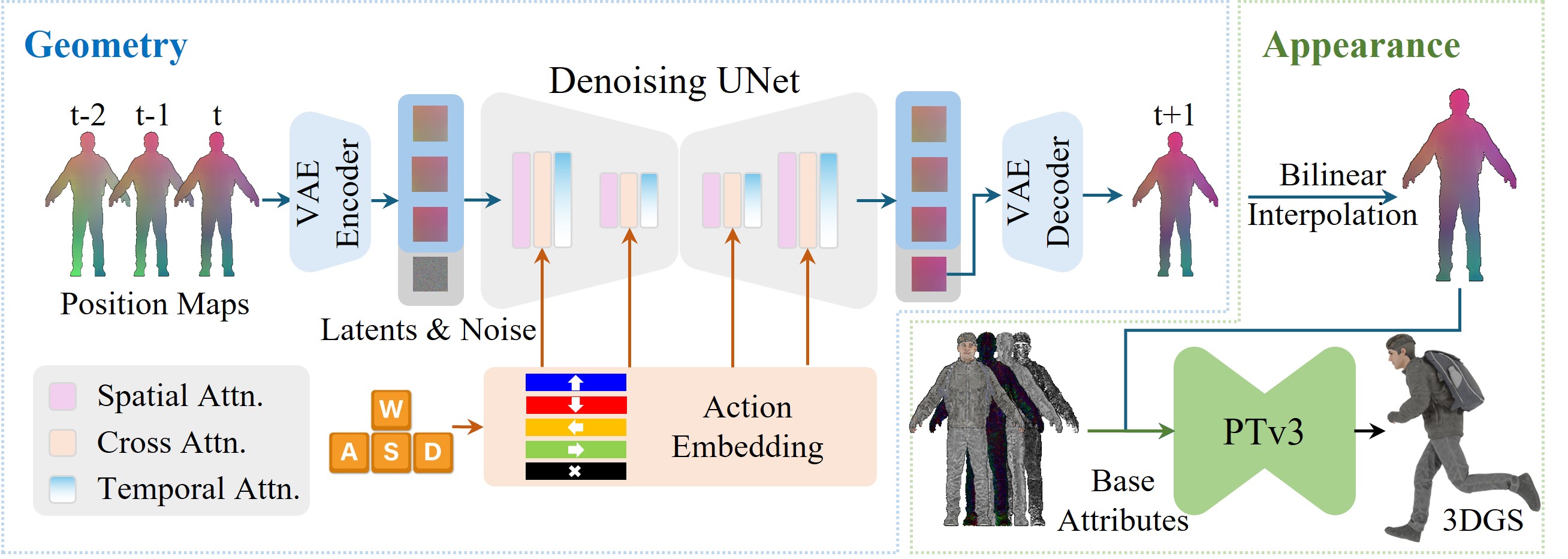}
    \vspace{-0.15in}
	\caption{The pipeline of \methodName{}. \methodName{} consists of an action-conditioned multi-frame diffusion model for avatar geometry and a point transformer for point-to-3DGS appearance mapping. The diffusion model takes latents of three position maps as context and generates one subsequent position map conditioned on a chosen action embedding. The generated position map is upscaled using bilinear interpolation and combined with other base 3DGS attributes to form a coarse 3DGS. The point transformer then refines the coarse 3DGS into one with avatar-specific appearance details. 
    }
    \vspace{-0.15in}
    \label{fig:pipeline}
\end{figure*}

\subsection{Task Formulation}
\label{sec:task_formulation}

In this section, we first formulate the action-driven neural avatar task. Let \(\mathbf{A}_n\) denote the 3D avatar at time \(n\), and let \(a_n\) denote a discrete action from the action space. Given a temporal context of previous avatar states $\mathbf{A}_{i \leq n}$ and the current action \(a_n\), the task is to generate one subsequent frame of the 3D avatar in updated dynamics:
\begin{equation}
    \mathbf{A}_{n+1} = f\bigl(\mathbf{A}_{i \leq n}; a_n\bigr),
    \label{eq:task_definition}
\end{equation}
where \(f\) represents the neural network. The key challenges are that the geometric transition from \(\mathbf{A}_n\) to $\mathbf{A}_{n+1}$ must exhibit physical plausibility, and $\mathbf{A}_{n+1}$ must possess a \(360^\circ\)-renderable appearance.

\subsubsection{Data Acquisition.}
Having defined the task, we now describe the data used to facilitate action-driven neural avatar training. We select five representative character models from the Unreal Engine Marketplace, encompassing non-rigid components such as robes and cloaks. For each avatar, we export its mesh sequences under five distinct key presses (Forward [W], Left [A], Backward [S], Right [D], and Idle [\(\emptyset\)]) as well as transitions between different actions, such as Forward to Backward ([W]\(\to\)[S]). Each frame in the motion sequence consists of a textured mesh of the avatar in a specific pose, paired with its corresponding action label. We use Unreal Engine's Chaos Cloth physics simulation system for clothing dynamics during movement. Details are provided in the supplementary material.

\subsection{Pipeline Overview}
The pipeline of \methodName{} is illustrated in \cref{fig:pipeline}. For the action-driven neural avatar task, we propose an action-conditioned multi-frame diffusion model for 4D geometry generation. For compatibility with well-established 2D diffusion architectures~\cite{ldm}, we encode 3D avatar geometry into 2D position maps. Temporal context $\mathbf{A}_{i \leq n}$ and the action signal $a_n$ are provided through different pathways. To synthesize temporally coherent geometric transitions, the diffusion model requires not only the current geometry but also velocity and acceleration information~\cite{vista}. Therefore, three historical context frames (\(t{-}2\) through \(t\)) are concatenated along the frame dimension and passed to the diffusion model. The fourth frame ($t+1$), which is to be generated, is initialized with Gaussian noise and gradually denoised through diffusion sampling. Discrete action signals are represented as learnable embeddings $\{e_k\}_{k=1}^{N}$, i.e., $N$ vectors of dimension $C$, where $N$ denotes the size of the action space ($N = 5$ in this work). These action embeddings modulate the generation process through cross attention.

Meanwhile, we follow a low-resolution geometry and high-resolution appearance paradigm to balance computational efficiency with rendering quality. The diffusion model operates on \(128 \times 128\) position maps to generate avatar geometry, while the point transformer produces dense 3DGS representations from upscaled \(1024 \times 1024\) position maps for high-fidelity appearance. This design decouples the computationally intensive generative modeling from appearance synthesis, enabling efficient action-conditioned generation without sacrificing visual detail.

\subsection{4D Geometry Representation}
\label{sec:representation}

\subsubsection{Position Map Rendering.}

\begin{figure*}[tb]
	\centering
	\includegraphics[width=0.95\textwidth, trim={0.75cm 0 0 0}, clip]{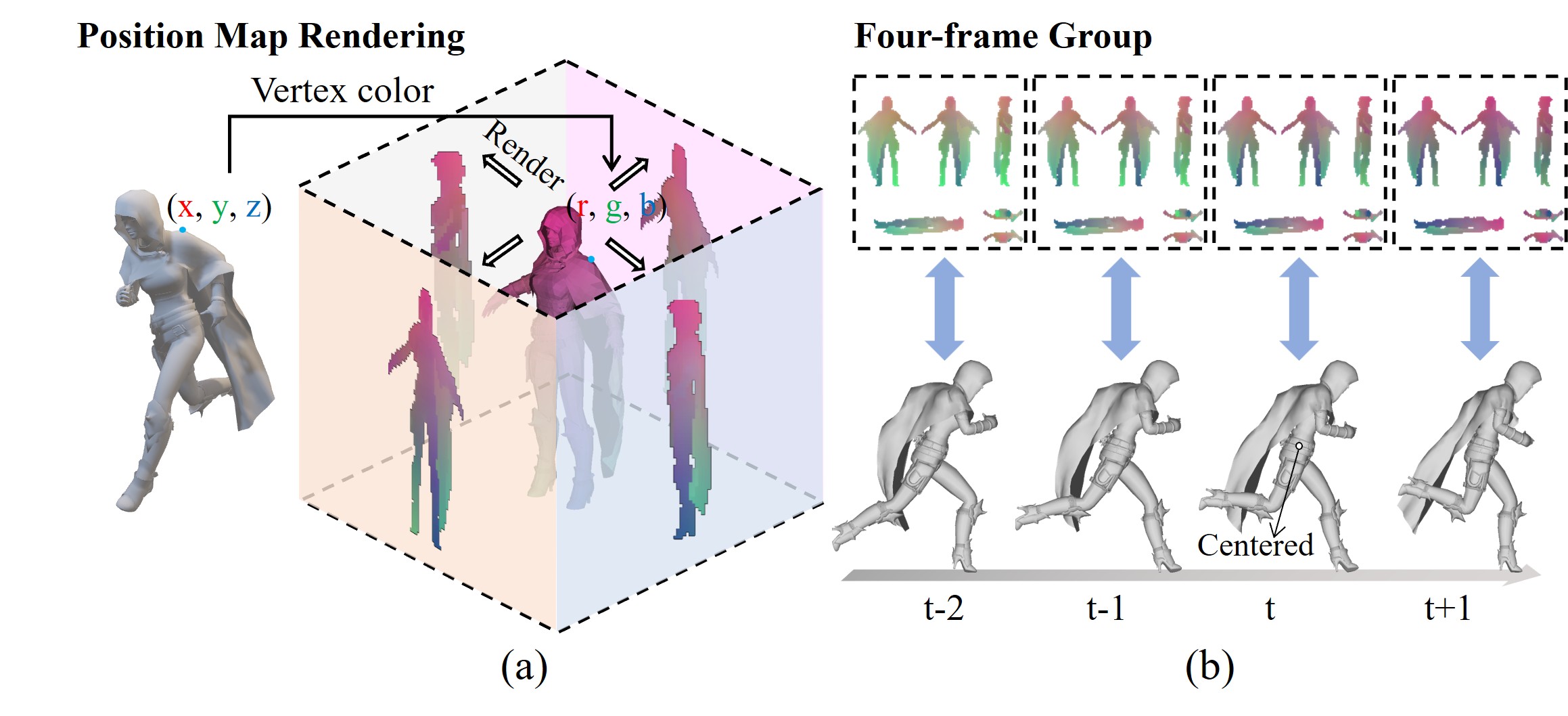}
    \vspace{-0.1in}
    \caption{The position map rendering process and visualization of a four-frame group. (a) We use an A-Pose mesh of the avatar as geometry and the vertex coordinates of the posed avatar as vertex colors to render position maps. The position maps are rendered from six orthogonal views. (b) We partition the motion sequence into groups of four frames and normalize each group.}
    \vspace{-0.15in}
    \label{fig:posmap_render}
\end{figure*}

The position map acquisition process is illustrated in \cref{fig:posmap_render}(a). Inspired by Animatable Gaussians~\cite{animatable_gaussians}, we employ a static avatar model in A-Pose as the reference mesh. The vertex coordinates of the avatar in a specific pose are mapped isometrically to the vertex colors of the corresponding vertices on this reference mesh. To ensure comprehensive body coverage, we render the reference mesh from six orthogonal views. On the rendered position map, the RGB value \((r, g, b)\) of each foreground pixel directly encodes a 3D point:
\begin{equation}
	\mathbf{P}(u, v) = (r, g, b) = (x, y, z),
	\label{eq:rgb_to_xyz}
\end{equation}
where \((u,v)\) are the pixel coordinates on the rendered position map \(\textbf{P}\) and \((x, y, z)\) denotes the 3D coordinates of the recovered point. We organize the six views into a single position map of size \(128 \times 128\), which compactly encodes the full geometry of the posed mesh.

\subsubsection{Group Normalization.}
A practical challenge arises from the fact that posed meshes in motion sequences have unbounded coordinates, analogous to an avatar freely exploring an open 3D world. However, neural network inputs and outputs must be constrained to a bounded range such as \([0, 1]\). To convert unbounded 3D coordinates to bounded pixel values,  we partition the motion sequences into overlapping temporal groups of four consecutive frames before position map rendering, as three context frames and one next frame are required. Consecutive groups are formed with a stride of one: the oldest frame is discarded and the next frame in the sequence is appended, so that adjacent groups share three frames. This sliding-window grouping also mirrors the autoregressive inference procedure (\cref{sec:inference}). 

For all groups across all motion sequences, we apply a uniform scaling factor \(s\) that normalizes the longest axis of the A-Posed reference mesh to 0.7. We then center the scaled third frame, i.e., frame $t$, in each group by translating its root to the origin and shifting the positions by 0.5:
\begin{equation}
	\mathbf{v}'_i = s \cdot \mathbf{v}_i - \mathbf{T}_t + \mathbf{0.5},
	\label{eq:normalization}
\end{equation}
where \(\mathbf{v}_i \in \mathbb{R}^3\) denotes the original vertex position, \(\mathbf{v}'_i\) is the normalized position, and \(\mathbf{T}_t\) is the translation vector that moves the root of frame \(t\) to the origin. The root is defined as the center of the avatar's waist. The same \(\mathbf{T}_t\) is applied to all other frames within the group. In this way, we normalize each four-frame group rather than individual meshes, allowing pixel changes from frame $t$ to $t+1$ to reflect both global root shift and local dynamics. The same $s$ is applied to all frames in all groups, guaranteeing consistent scaling of avatar geometry.
Meanwhile, since the longest axis of the reference mesh is scaled to \(0.7\) rather than the full \([0.0, 1.0]\) range, a sufficient margin is reserved to accommodate the displacement of neighboring frames, which are at most two frames away from the centered frame \(t\). This ensures that all normalized vertex coordinates remain within \([0.0, 1.0]\) and can be directly mapped to pixel values. 

In \cref{fig:posmap_render}(b), we visualize position maps and their corresponding meshes within a group. The global variation in hue reflects the forward motion of the avatar, while local variations accurately capture limb and clothing dynamics. For each avatar, the 4-frame groups serve as direct training samples for the multi-frame diffusion model, which learns physically plausible geometric deformations encoded in the position map groups. The ground-truth action label for each group is the label of frame \(t\), which drives the avatar to frame \(t+1\).

\subsection{Action-Conditioned Geometry Generation}
\label{sec:button_control}

The diffusion model in \methodName{} generates action-conditioned avatar geometry via a UNet-based latent diffusion architecture~\cite{unet, ddpm}. Frames \(t{-}2\) through \(t\) are encoded into latent representations using an avatar-specific Variational AutoEncoder (VAE) and concatenated along the frame dimension. Let \(\mathbf{z}_i = \mathcal{E}(\mathbf{P}_i)\) denote the latent encoding of position map \(\mathbf{P}_i\), where \(\mathcal{E}\) represents the VAE encoder. The next frame to be generated is initialized with Gaussian noise \(\boldsymbol{\epsilon} \sim \mathcal{N}(\mathbf{0}, \mathbf{I})\) and concatenated likewise, yielding the input tensor:
\begin{equation}
	\mathbf{Z} = \left[\mathbf{z}_{t-2}, \mathbf{z}_{t-1}, \mathbf{z}_t, \mathbf{z}_{t+1}^{(\tau)}\right] \in \mathbb{R}^{4 \times c \times h \times w},
	\label{eq:latent_concat}
\end{equation}
where \(\mathbf{z}_{t+1}^{(\tau)}\) denotes the noisy latent at diffusion timestep \(\tau\), and \(c\), \(h\), \(w\) represent the channel, height, and width dimensions of the latent space, respectively.

The denoising UNet $\boldsymbol{\epsilon}_\theta$ is trained to predict the noise $\boldsymbol{\epsilon}$ added to $\mathbf{z}_{t+1}$, conditioned on the concatenated input tensor $\mathbf{Z}$ and the action embedding $e_k$. The conditional diffusion training objective is:
\begin{equation}
	\mathcal{L}_{\text{diff}} = \mathbb{E}_{\mathbf{z}_{t+1},\, \boldsymbol{\epsilon} \sim \mathcal{N}(\mathbf{0}, \mathbf{I}),\, \tau} \left[ \left\| \boldsymbol{\epsilon} - \boldsymbol{\epsilon}_\theta\!\left(\mathbf{Z},\, \tau,\, e_k\right) \right\|_2^2 \right],
	\label{eq:diffusion_objective}
\end{equation}
where $\tau$ denotes the diffusion timestep. The network incorporates convolutional layers enhanced with spatial attention, cross attention, and temporal attention modules stacked sequentially. \textit{Spatial attention} operates within each individual frame. Since we organize position maps from six viewpoints into a single image, standard spatial attention suffices to establish cross-view connections and enhance multi-view consistency, eliminating the need for explicit multi-view attention modules. \textit{Cross attention} is where the action condition takes effect. During inference, the appropriate action embedding $e_k$ is selected according to user input, thereby controlling the generation of the next frame. \textit{Temporal attention} is computed across the same spatial positions of all four frames, transferring information about the character's current position, velocity, and acceleration from frames \(t{-}2\) through \(t\) into the generation of frame \(t{+}1\). Since all position maps share the same spatial layout, temporal attention inherently operates on corresponding body parts across frames. The denoising UNet outputs latents for all four frames, of which only the fourth is passed to the VAE decoder \(\mathcal{D}\) to produce the generated position map \(\hat{\mathbf{P}}_{t+1} = \mathcal{D}(\hat{\mathbf{z}}_{t+1})\).

\subsection{Appearance Mapping}
\label{sec:appearance}

For a given avatar identity, appearance is a function of geometry, since the same surface point carries the same texture regardless of pose. As position maps can be directly converted to point clouds, we employ a modified Point Transformer V3 model (PTv3)~\cite{ptv3, splatformer} to map geometry to appearance represented as 3DGS.

To improve efficiency, the generated position map is produced at a low resolution. If converted to a point cloud with a one-to-one pixel-to-point mapping, the resulting point count would be insufficient for detailed avatar appearance. Therefore, we first upscale the position map from \(128 \times 128\) to \(1024 \times 1024\) using foreground-aware bilinear interpolation that prevents blending valid foreground coordinates with zero-valued background pixels. 

To facilitate the appearance mapping task, we employ a coarse-to-fine strategy. Specifically, we combine the upscaled position map with a base 3DGS attribute map \(\mathcal{A}_{\text{base}}\), trained on a single frame of the avatar in a standing pose, to form a coarse 3DGS. While spherical harmonics, scaling, rotation, and opacity of the coarse 3DGS come from \(\mathcal{A}_{\text{base}}\), Gaussian means come from the upscaled position map. Then, we pass the coarse 3DGS to PTv3, which predicts pose-dependent attribute offsets \(\Delta \mathcal{G}\):
\begin{equation}
	\mathcal{G}_{\text{refined}} = \mathcal{G}_{\text{coarse}} + \Delta \mathcal{G},
	\label{eq:3dgs_refine}
\end{equation}
where \(\mathcal{G}_{\text{coarse}}\) denotes the coarse 3DGS attributes and \(\mathcal{G}_{\text{refined}}\) represents the refined attributes. The refined 3DGS supports high-fidelity \(360^\circ\) rendering.

\subsection{Training Strategy}
\label{sec:training}

We first train a VAE on the position maps of each avatar using a combination of reconstruction loss and KL divergence loss. We then divide diffusion model training into two stages. In the first stage, we jointly train the five learnable action embeddings $\{e_k\}_{k=1}^{5}$ and the UNet exclusively on training groups with consistent movement directions, allowing the action embeddings to develop distinct directional semantics. In the second stage, we freeze the action embeddings, incorporate the turning frames with oversampling, and train only the UNet to handle directional transitions. Classifier-Free Guidance (CFG)~\cite{cfg} is employed in both stages for conditional generation. In order to improve robustness during autoregressive inference, following prior work~\cite{gamengen, diffusion_forcing}, we also inject independent noise into the context frames during training. The diffusion loss in \cref{eq:diffusion_objective} is computed exclusively on the fourth frame, while the three context frames serve solely as conditioning input. Finally, PTv3 is trained on renderings across all poses to learn appearance refinements. Further details are provided in the supplementary material.

\subsection{Progressive 4D Inference}
\label{sec:inference}

\begin{figure*}[tb]
	\centering
	\includegraphics[width=\textwidth]{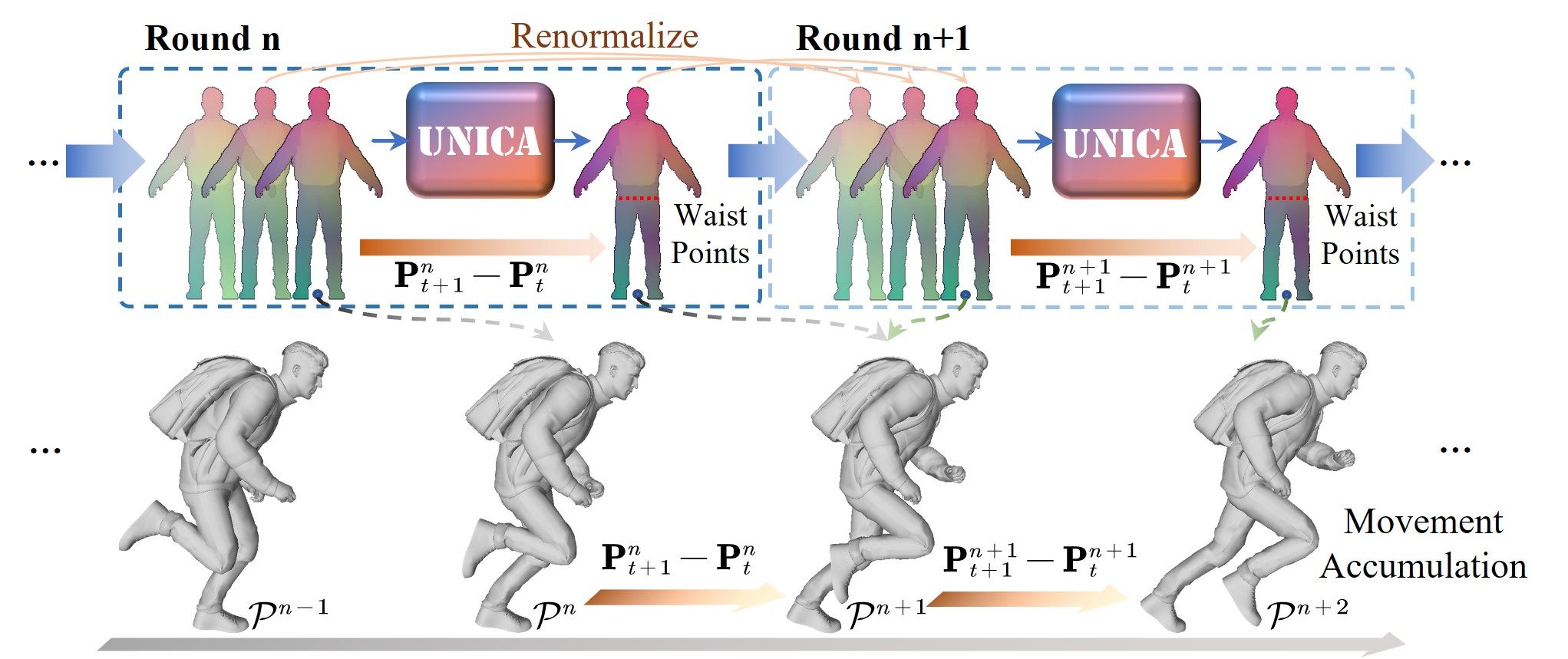}
    \vspace{-0.2in}
    \caption{Demonstration of progressive 4D inference. During autoregressive inference of \methodName{}, each round generates a relative movement that is accumulated in 3D space for the actual movement of the 3D avatar. The output frame of round \(n\) will be renormalized before it is used as input for round \(n+1\).}
    \vspace{-0.15in}
	\label{fig:inference}
\end{figure*}

Single-frame generation in the diffusion model follows the standard DDIM~\cite{ddim} sampling procedure. Following Animatable Gaussians~\cite{animatable_gaussians}, we apply Principal Component Analysis (PCA) fitted on the training set to reconstruct the generated position map, aligning the six views in 3D coordinates. 

To enable avatar navigation in an unbounded 3D world, we propose the progressive 4D inference strategy during autoregressive inference of \methodName{}, divided into movement accumulation and renormalization, as shown in \cref{fig:inference}. Movement accumulation computes the character's world-coordinate geometry by progressively accumulating relative increments. Let \(\mathcal{P}^n\) denote the avatar's world-space geometry positions at time \(n\). The accumulated positions are obtained with:
\begin{equation}
	\mathcal{P}^{n+1} = \mathcal{P}^n + \left(\mathbf{P}_{t+1}^{n} - \mathbf{P}_t^{n}\right),
	\label{eq:motion_accum}
\end{equation}
where \(\mathbf{P}_t^{n}\) and \(\mathbf{P}_{t+1}^{n}\) denote positions of frames \(t\) and \(t{+}1\) in the normalized coordinate space of round \(n\). This is enabled by our group normalization design, which ensures that all four frames within a group share the same coordinate space, so that the difference \(\mathbf{P}_{t+1}^{n} - \mathbf{P}_t^{n}\) represents real movement.

Renormalization is necessary to ensure pixel values remain within the valid range. We compute the root position of \(\mathbf{P}_{t+1}^{n}\) by averaging its waist points on the position map, then translate the root to the origin, so that the renormalized \(\mathbf{P}_{t+1}^{n}\) serves as frame $t$ in the next round:
\begin{equation}
	\mathbf{P}_{t}^{n+1} = \mathbf{P}_{t+1}^{n} - \mathbf{T}_{t+1}^{n},
	\label{eq:renorm}
\end{equation}
where \(\mathbf{T}_{t+1}^{n}\) is the translation vector. The same translation is applied to preceding frames, transforming them into the new context for round \(n{+}1\). Notably, renormalization operates on the position maps prior to PCA reconstruction, ensuring that PCA does not influence the accumulated motion sequence.

On the appearance end, PTv3 operates in normalized space and modifies the position attribute of 3DGS. We thus apply Procrustes analysis to recover the optimal transformations that align the PTv3 output to world coordinates without distorting the surface geometry.
\section{Experiment}
\label{sec:experiment}

\subsection{Animation Results}

\begin{figure*}[t]
	\centering
	\includegraphics[width=\textwidth]{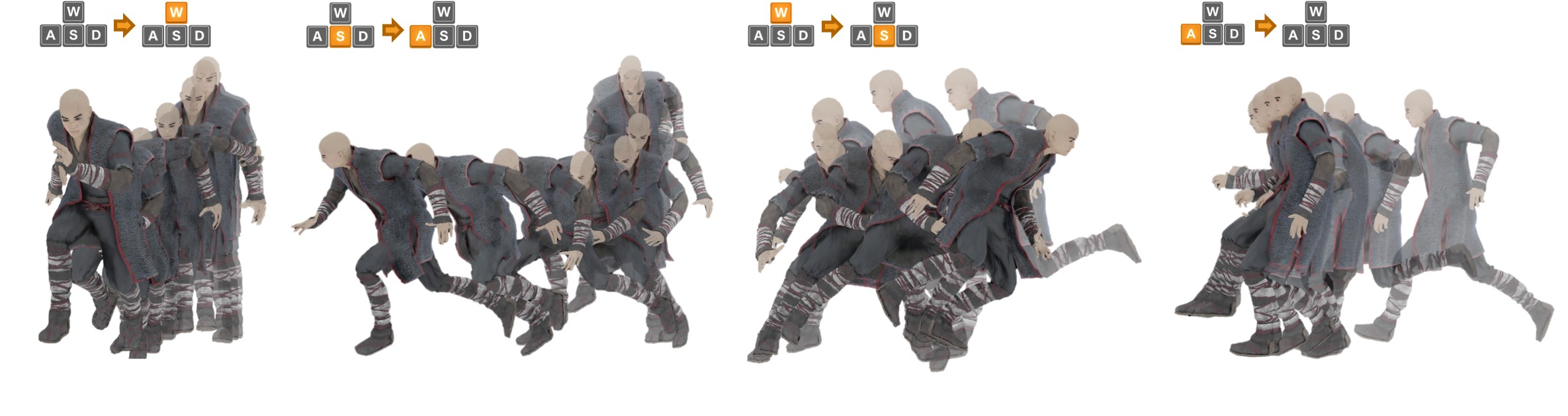}
    \vspace{-0.2in}
    \caption{Animation results of \methodName{} demonstrating avatar response to key presses. For visualization clarity, we sample one frame every three frames along the trajectory.}
    \vspace{-0.05in}
	\label{fig:ani_results}
\end{figure*}

\subsubsection{Implementation Details.} All models are trained on 8 NVIDIA A6000 GPUs. The VAE and denoising UNet are each trained for 300k iterations. Based on experimental comparison, we adopt 10-step DDIM sampling during inference. PTv3 is trained using 50 views rendered along a cylindrical surface for 20 epochs. Additional details are provided in the supplementary material. 

\subsubsection{Results.} As \methodName{} introduces a new task — action-driven neural avatar — we first present its basic functionality in \cref{fig:ani_results}: avatars responding to key presses through autoregressive inference. The figures are rendered by compositing 3DGS frames into a single scene without manual manipulation, demonstrating that the avatar genuinely moves through 3D space via our progressive 4D inference technique. Physically plausible swaying of robe hems is also inherently captured. 

\begin{figure*}[t]
	\centering
	\includegraphics[width=0.75\textwidth]{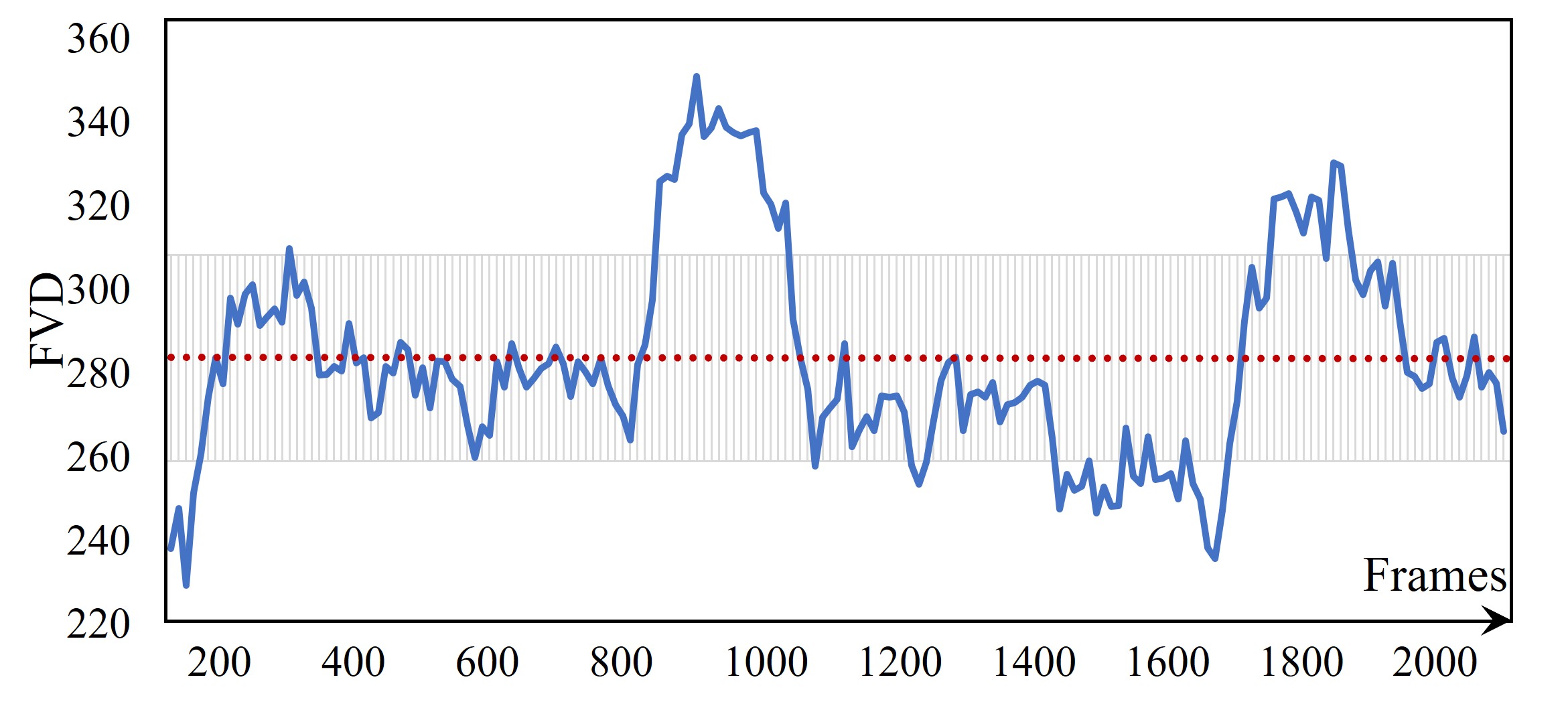}
    \vspace{-0.1in}
    \caption{FVD scores over a 2000-frame autoregressively generated sequence, computed using a sliding window of 200 frames with a stride of 10. The red dashed line indicates the linear trend, and the gray shaded region denotes $\pm$1 standard deviation. The near-flat trend demonstrates stable generation quality over extended rollouts.}
    \vspace{-0.1in}
	\label{fig:auto_regressive}
\end{figure*}

We further verify extra-long duration generation by repeating an action sequence \{60$\times$[W], 60$\times$[S], 60$\times$[A], 20$\times$[$\emptyset$]\} ten times, generating 2,000 frames per avatar. \Cref{fig:auto_regressive} reports Fréchet Video Distance (FVD) computed with a 200-frame sliding window. The linear trend shows that FVD fluctuates around the mean without upward drift, confirming that \methodName{} maintains stable quality over extended rollouts. We further evaluate control responsiveness: the generated sequences require on average 13.4 and 9.8 frames for $180^\circ$ and $90^\circ$ turns, respectively, closely matching the ground-truth values of 12 and 9. Given that the avatar cannot initiate a turn while both feet are off the ground, these statistics confirm that \methodName{} responds promptly to changes in action inputs.

\subsection{Comparisons}

\subsubsection{Metrics.} 
Following~\cite{gamengen}, we evaluate at both image and video levels. For image-level evaluation, we generate a single frame conditioned on three input frames and the ground-truth action label, then compare against the ground-truth fourth frame using Peak Signal-to-Noise Ratio (PSNR) and Learned Perceptual Image Patch Similarity (LPIPS) across four cardinal viewpoints. For video-level evaluation, we autoregressively generate a 200-frame trajectory comprising Forward, Backward, Left, and Idle segments, encompassing one start, one $180^\circ$ turn, one $90^\circ$ turn, and one stop. We compute FVD~\cite{fvd} between rendered videos from four cardinal viewpoints and ground-truth references containing equivalent action events and the same number of frames. All reported metrics are averaged across the four viewpoints.

\subsubsection{Baselines.} 
We compare against Animatable Gaussians~\cite{animatable_gaussians}, a state-of-the-art animatable 3D avatar method, and Mixamo~\cite{mixamo}, a widely adopted automatic rigging tool. 
Since these baselines are not end-to-end solutions like ours, we separately provide them with ground-truth inputs: SMPL-X parameters for Animatable Gaussians and skeleton sequences for Mixamo.
We additionally include SV4D 2.0~\cite{sv4d20} for visual comparison. Baseline implementation details are provided in the supplementary material.

\subsubsection{Qualitative Results.} 

\begin{figure*}[t]
	\centering
	\includegraphics[width=\textwidth]{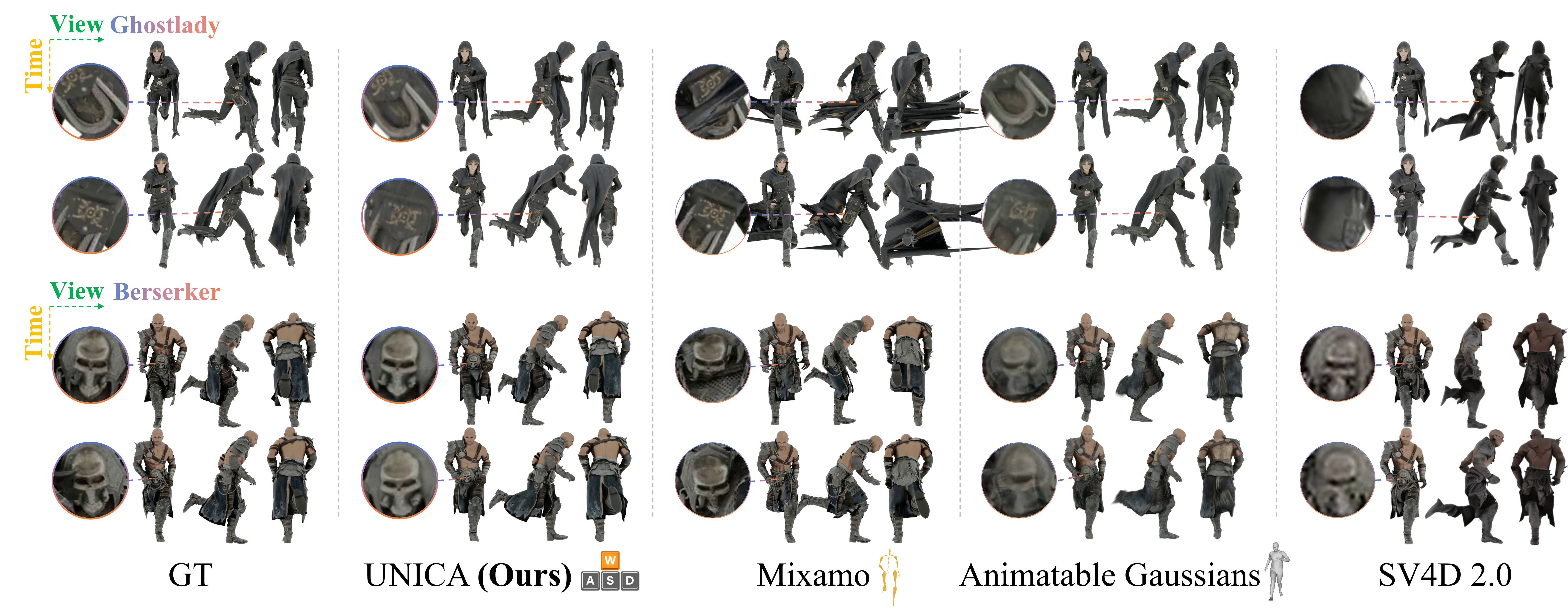}
    \vspace{-0.15in}
    \caption{Qualitative comparison between \methodName{} and baseline methods across different frames and viewpoints.}
    \vspace{-0.15in}
	\label{fig:vis_comp}
\end{figure*}

\Cref{fig:vis_comp} presents visual comparisons. All \methodName{} results are obtained through autoregressive inference, reflecting practical rendering quality. \methodName{} produces high-quality avatars closely matching ground truth, with strong performance on faces, bodies, and textural details, while generating natural dynamics for loose clothing.

While Mixamo achieves high rendering quality through ground-truth mesh and texture mapping, automatic rigging fails to model loose garments: for \textit{Ghostlady}, the cloak binds erroneously to the body skeleton, causing severe deformation; for \textit{Berserker}, the hem adheres unnaturally to the legs. Animatable Gaussians can model non-rigid components but exhibits noticeable artifacts such as distorted leg geometry of \textit{Berserker} and reduced texture fidelity on intricate clothing elements. SV4D 2.0 generalizes across different cases but does not achieve quality comparable to avatar-specific methods.

\subsubsection{Quantitative Results and User Study.} 

\begin{table}[t]
  \caption{Quantitative comparison and user study results averaged across five avatars. User study ratings are on a 1--5 scale. Best results in \textbf{bold}. VQS~$=$ Visual quality score in user study; PRS~$=$ Physical realism score in user study.}
  \vspace{-0.1in}
  \label{tab:comparison}
  \centering
  \begin{tabularx}{\textwidth}{lYYY|YY}
    \toprule
    Method & FVD $\downarrow$ & PSNR $\uparrow$ & LPIPS $\downarrow$ & VQS $\uparrow$ & PRS $\uparrow$ \\
    \midrule
    Ani. Gaussians~\cite{animatable_gaussians} & 199.93 & 20.31 & 0.119 & 3.47 & 3.64 \\
    Mixamo~\cite{mixamo} & 320.06 & 14.54 & 0.187 & 3.39 & 2.64 \\
    \methodName{} (Ours) & \textbf{194.87} & \textbf{23.95} & \textbf{0.106} & \textbf{4.36} & \textbf{4.24} \\
    \bottomrule
  \end{tabularx}
  \vspace{-0.05in}
\end{table}

\Cref{tab:comparison} reports quantitative metrics and user study results averaged across all five avatars. We asked participants to rate Visual Quality Score (VQS) and Physical Realism Score (PRS) for side-by-side videos with ground truth as reference. \methodName{} outperforms baselines on all metrics. Note that the metrics of Animatable Gaussians reflect training-pose performance, whereas \methodName{} is evaluated through autoregressive inference on novel trajectories.

\subsection{Ablation Study}

\begin{figure*}[t]
	\centering
	\includegraphics[width=0.9\textwidth]{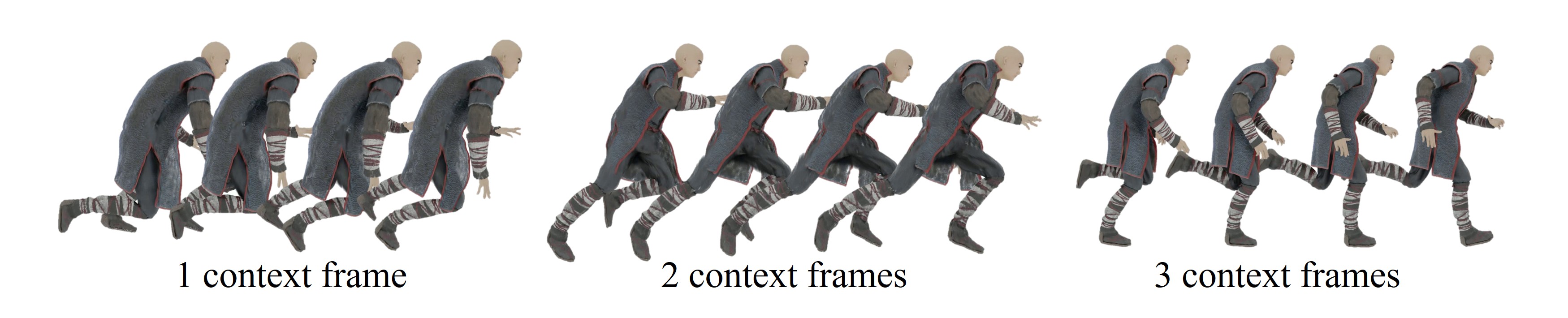}
    \vspace{-0.15in}
	\caption{Comparison on the number of context frames. Four consecutive frames are shown for each method.}
    \vspace{-0.15in}
	\label{fig:context_frames}
\end{figure*}

\subsubsection{Minimal Context Frames.}
Three consecutive frames constitute the minimum context required to capture second-order motion dynamics (position, velocity, and acceleration). \Cref{fig:context_frames} empirically validates this: reducing the context to one or two frames noticeably degrades motion quality, with the avatar exhibiting unnatural bending poses and disordered limb movements. With three context frames, \methodName{} produces smooth and coherent avatar motion while keeping computation and memory overhead minimal.

\begin{table}[t]
\centering
\caption{Comparison of results with and without PCA alignment.}
\label{tab:pca}
\vspace{-0.1in}
\begin{tabularx}{0.55\textwidth}{YYYY}
\toprule
 & FVD $\downarrow$ & PSNR $\uparrow$ & LPIPS $\downarrow$ \\
\midrule
w/o PCA & 228.31 & 23.68 & 0.114 \\
w/ PCA & \textbf{194.87} & \textbf{23.95} & \textbf{0.106} \\
\bottomrule
\end{tabularx}
\vspace{-0.05in}
\end{table}

\begin{figure*}[t]
	\centering
	\includegraphics[width=0.8\textwidth]{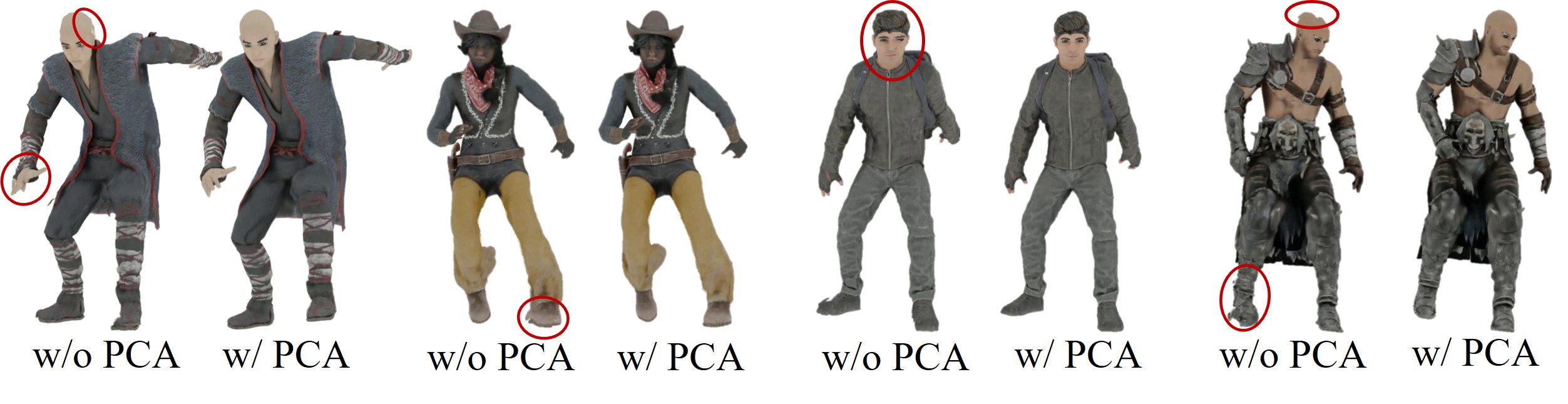}
    \vspace{-0.1in}
	\caption{Ablation study of position map alignment via PCA reconstruction.}
    \vspace{-0.15in}
	\label{fig:pca}
\end{figure*}

\subsubsection{Position Map Alignment.}
Each generated frame of avatar geometry comprises position maps from six views. Although spatial attention provides global reasoning for multi-view consistency, generation inaccuracies can cause misalignment. As shown in \cref{fig:pca}, passing raw geometry directly to PTv3 produces fragmented artifacts at challenging turning poses—heads, faces, and shoes appear broken due to noisy, misaligned pixels. PCA reconstruction aligns the six views with minimal pose accuracy loss by projecting onto principal components that inherently encode multi-view consistency, substantially improving avatar appearance. Quantitative results are shown in \cref{tab:pca}.

\subsection{Limitations and Future Work}
While \methodName{} demonstrates the feasibility of action-driven 3D avatars through neural networks, several limitations remain. First, the current inference time leaves a gap before real-time deployment. Second, the 3DGS representations in certain turning poses exhibit artifacts. Details are presented in the supplementary material.  SSecond, \methodName{} currently supports only locomotion; extending to a larger action space is a promising direction. 
\section{Conclusion}
\label{sec:conclusion}

We have presented \methodName{}, a framework that brings video-game-style control to 3D avatars. Given discrete keyboard inputs, \methodName{} generates full-body avatar motion with realistic appearance and physically plausible dynamics—without skeleton rigs, motion planning, or explicit physical simulation. Our skeleton-free pipeline combines action-conditioned diffusion on position maps with a point transformer outputting 3D Gaussian Splatting. Through group normalization and progressive 4D inference, \methodName{} learns unbounded locomotion from mesh sequences and maintains stable quality in autoregressive generation. This work opens new possibilities for creating controllable avatars from light-stage captures.

%
%
\bibliographystyle{splncs04}
\bibliography{main}

@String(CVPR  = {IEEE Conf. Comput. Vis. Pattern Recog.})

@String(ICCV  = {Int. Conf. Comput. Vis.})

@String(ECCV  = {Eur. Conf. Comput. Vis.})

@String(NeurIPS = {Adv. Neural Inform. Process. Syst.})

@String(ICML  = {Int. Conf. Mach. Learn.})

@String(ICLR  = {Int. Conf. Learn. Represent.})

@String(ICME  = {Int. Conf. Multimedia and Expo})

@String(TOG   = {ACM Trans. Graph.})

@String(TIP   = {IEEE Trans. Image Process.})

@String(CVPR  = {CVPR})

@String(ICCV  = {ICCV})

@String(ECCV  = {ECCV})

@String(NeurIPS = {NeurIPS})

@String(ICML  = {ICML})

@String(ICLR  = {ICLR})

@String(ICME  =	{ICME})

@String(TOG   = {ACM TOG})

@String(TIP   = {IEEE TIP})

@String(THREEDV = {3DV})

@String(SIGGRAPHASIA = {SIGGRAPH Asia})

@String(SIGGRAPH = {SIGGRAPH})

@String(Nature = {Nature})

@String(MICCAI = {MICCAI})

@String(NeurIPSW = {NeurIPS Workshop})

@String(CSUR = {ACM Comput. Surv.})

@inproceedings{pred_animation_skeletons,
  author    = {Xu, Zhan and Zhou, Yang and Kalogerakis, Evangelos and Singh, Karan},
  booktitle = THREEDV,
  title     = {Predicting Animation Skeletons for {3D} Articulated Models via Volumetric Nets},
  year      = {2019},
  pages     = {298-307},
}

@article{riggnet,
  author    = {Xu, Zhan and Zhou, Yang and Kalogerakis, Evangelos and Landreth, Chris and Singh, Karan},
  title     = {{RigNet}: Neural Rigging for Articulated Characters},
  year      = {2020},
  volume    = {39},
  number    = {4},
  journal   = TOG,
  pages     = {58:1--58:14}
}

@article{unirig,
  author    = {Zhang, Jia-Peng and Pu, Cheng-Feng and Guo, Meng-Hao and Cao, Yan-Pei and Hu, Shi-Min},
  title     = {One Model to Rig Them All: Diverse Skeleton Rigging with UniRig},
  year      = {2025},
  volume    = {44},
  number    = {4},
  journal   = TOG,
  pages     = {123:1--123:18}
}

@article{neural_blend_shape,
  author    = {Li, Peizhuo and Aberman, Kfir and Hanocka, Rana and Liu, Libin and Sorkine-Hornung, Olga and Chen, Baoquan},
  title     = {Learning Skeletal Articulations with Neural Blend Shapes},
  year      = {2021},
  volume    = {40},
  number    = {4},
  journal   = TOG,
  pages     = {130:1--130:15}
}

@inproceedings{skinningnet,
  author    = {Mosella-Montoro, Albert and Ruiz-Hidalgo, Javier},
  title     = {{SkinningNet}: Two-Stream Graph Convolutional Neural Network for Skinning Prediction of Synthetic Characters},
  booktitle = CVPR,
  year      = {2022},
  pages     = {18593-18602}
}

@article{riganything,
  author    = {Liu, Isabella and Xu, Zhan and Yifan, Wang and Tan, Hao and Xu, Zexiang and Wang, Xiaolong and Su, Hao and Shi, Zifan},
  title     = {{RigAnything}: Template-Free Autoregressive Rigging for Diverse {3D} Assets},
  year      = {2025},
  volume    = {44},
  number    = {4},
  journal   = TOG,
  pages     = {122:1--122:12}
}

@article{pfnn,
  author    = {Holden, Daniel and Komura, Taku and Saito, Jun},
  title     = {Phase-Functioned Neural Networks for Character Control},
  year      = {2017},
  volume    = {36},
  number    = {4},
  journal   = TOG,
  pages     = {42:1--42:13}
}

@article{mode_adaptive_nn,
  author    = {Zhang, He and Starke, Sebastian and Komura, Taku and Saito, Jun},
  title     = {Mode-Adaptive Neural Networks for Quadruped Motion Control},
  year      = {2018},
  volume    = {37},
  number    = {4},
  journal   = TOG,
  pages     = {145:1--145:11}
}

@article{neural_state_machine,
  author  = {Starke, Sebastian and Zhang, He and Komura, Taku and Saito, Jun},
  title   = {Neural State Machine for Character-Scene Interactions},
  year    = {2019},
  volume  = {38},
  number  = {6},
  journal = TOG,
  pages   = {209:1--209:14}
}

@article{local_motion_phases,
  author    = {Starke, Sebastian and Zhao, Yiwei and Komura, Taku and Zaman, Kazi},
  title     = {Local Motion Phases for Learning Multi-Contact Character Movements},
  year      = {2020},
  volume    = {39},
  number    = {4},
  journal   = TOG,
  pages     = {54:1--54:13}
}

@article{deepphase,
  author    = {Starke, Sebastian and Mason, Ian and Komura, Taku},
  title     = {{DeepPhase}: Periodic Autoencoders for Learning Motion Phase Manifolds},
  year      = {2022},
  volume    = {41},
  number    = {4},
  journal   = TOG,
  pages     = {136:1--136:13}
}

@inproceedings{neural_motion_graph,
  author    = {Tao, Hongyu and Hou, Shuaiying and Zou, Changqing and Bao, Hujun and Xu, Weiwei},
  title     = {Neural Motion Graph},
  year      = {2023},
  booktitle = SIGGRAPHASIA,
  pages     = {84:1--84:11}
}

@article{smpl,
  author    = {Loper, Matthew and Mahmood, Naureen and Romero, Javier and Pons-Moll, Gerard and Black, Michael J.},
  title     = {{SMPL}: A Skinned Multi-Person Linear Model},
  journal   = TOG,
  volume    = {34},
  number    = {6},
  pages     = {248:1--248:16},
  year      = {2015},
}

@article{smplh,
  author    = {Romero, Javier and Tzionas, Dimitrios and Black, Michael J.},
  title     = {Embodied Hands: Modeling and Capturing Hands and Bodies Together},
  journal   = TOG,
  volume    = {36},
  number    = {6},
  pages     = {245:1--245:17},
  year      = {2017},
}

@inproceedings{smplx,
  author    = {Pavlakos, Georgios and Choutas, Vasileios and Ghorbani, Nima and Bolkart, Timo and Osman, Ahmed A. A. and Tzionas, Dimitrios and Black, Michael J.},
  title     = {Expressive Body Capture: {3D} Hands, Face, and Body From a Single Image},
  booktitle = CVPR,
  year      = {2019},
  pages     = {10975--10985}
}

@inproceedings{animatable_gaussians,
  author    = {Li, Zhe and Zheng, Zerong and Wang, Lizhen and Liu, Yebin},
  title     = {{Animatable Gaussians}: Learning Pose-dependent {Gaussian} Maps for High-fidelity Human Avatar Modeling},
  booktitle = CVPR,
  pages     = {19711--19722},
  year      = {2024}
}

@inproceedings{smpl_based1,
  author    = {Zheng, Zerong and Huang, Han and Yu, Tao and Zhang, Hongwen and Guo, Yandong and Liu, Yebin},
  title     = {Structured Local Radiance Fields for Human Avatar Modeling},
  booktitle = CVPR,
  pages     = {15893--15903},
  year      = {2022}
}

@inproceedings{taoavatar,
  author    = {Chen, Jianchuan and Hu, Jingchuan and Wang, Gaige and Jiang, Zhonghua and Zhou, Tiansong and Chen, Zhiwen and Lv, Chengfei},
  title     = {{TaoAvatar}: Real-Time Lifelike Full-Body Talking Avatars for Augmented Reality via {3D Gaussian Splatting}},
  booktitle = CVPR,
  pages     = {10723--10734},
  year      = {2025}
}

@inproceedings{gaussian_avatar,
  author    = {Hu, Liangxiao and Zhang, Hongwen and Zhang, Yuxiang and Zhou, Boyao and Liu, Boning and Zhang, Shengping and Nie, Liqiang},
  title     = {{GaussianAvatar}: Towards Realistic Human Avatar Modeling from a Single Video via Animatable {3D Gaussians}},
  booktitle = CVPR,
  pages     = {634--644},
  year      = {2024}
}

@inproceedings{multi_hmr,
  author    = {Baradel, Fabien and Armando, Matthieu and Galaaoui, Salma and Br{\'e}gier, Romain and Weinzaepfel, Philippe and Rogez, Gr{\'e}gory and Lucas, Thomas},
  title     = {Multi-{HMR}: Multi-Person Whole-Body Human Mesh Recovery in a Single Shot},
  booktitle = ECCV,
  pages     = {202--218},
  year      = {2024}
}

@inproceedings{pixie,
  author    = {Feng, Yao and Choutas, Vasileios and Bolkart, Timo and Tzionas, Dimitrios and Black, Michael J.},
  title     = {Collaborative Regression of Expressive Bodies using Moderation},
  booktitle = THREEDV,
  pages     = {792--804},
  year      = {2021},
}

@inproceedings{osx,
  author    = {Lin, Jing and Zeng, Ailing and Wang, Haoqian and Zhang, Lei and Li, Yu},
  title     = {One-Stage {3D} Whole-Body Mesh Recovery with Component Aware Transformer},
  booktitle = CVPR,
  pages     = {21159--21168},
  year      = {2023}
}

@inproceedings{4dfy,
  author    = {Bahmani, Sherwin and Skorokhodov, Ivan and Rong, Victor and Wetzstein, Gordon and Guibas, Leonidas and Wonka, Peter and Tulyakov, Sergey and Park, Jeong Joon and Tagliasacchi, Andrea and Lindell, David B.},
  title     = {{4D-fy}: Text-to-{4D} Generation Using Hybrid Score Distillation Sampling},
  booktitle = CVPR,
  pages     = {7996--8006},
  year      = {2024}
}

@inproceedings{easymocap,
  title={Motion capture from internet videos},
  author={Dong, Junting and Shuai, Qing and Zhang, Yuanqing and Liu, Xian and Zhou, Xiaowei and Bao, Hujun},
  booktitle=ECCV,
  pages={210--227},
  year={2020},
}

@inproceedings{sv4d,
  author    = {Xie, Yiming and Yao, Chun-Han and Voleti, Vikram and Jiang, Huaizu and Jampani, Varun},
  title     = {{SV4D}: Dynamic {3D} Content Generation with Multi-Frame and Multi-View Consistency},
  booktitle = ICLR,
  year      = {2025}
}

@article{sv4d20,
  author    = {Yao, Chun-Han and Xie, Yiming and Voleti, Vikram and Jiang, Huaizu and Jampani, Varun},
  title     = {{SV4D} 2.0: Enhancing Spatio-Temporal Consistency in Multi-View Video Diffusion for High-Quality {4D} Generation},
  journal   = {arXiv preprint arXiv:2503.16396},
  year      = {2025}
}

@inproceedings{l4gm,
  author    = {Ren, Jiawei and Xie, Kevin and Mirzaei, Ashkan and others},
  title     = {{L4GM}: Large {4D Gaussian} Reconstruction Model},
  booktitle = NeurIPS,
  volume    = {37},
  pages     = {56828--56858},
  year      = {2024},
}

@article{world_model_review,
  author    = {Ding, Jingtao and Zhang, Yunke and Shang, Yu and others},
  title     = {Understanding World or Predicting Future? {A} Comprehensive Survey of World Models},
  journal   = CSUR,
  volume    = {58},
  number    = {3},
  articleno = {57},
  year      = {2025},
}

@article{lingbot_world,
  author    = {Gao, Zelin and Wang, Qiuyu and Zeng, Yanhong and others},
  title     = {Advancing Open-source World Models},
  journal   = {arXiv preprint arXiv:2601.20540},
  year      = {2026}
}

@article{matrix_game,
  author    = {Zhang, Yifan and Peng, Chunli and Wang, Boyang and others},
  title     = {{Matrix-Game}: Interactive World Foundation Model},
  journal   = {arXiv preprint arXiv:2506.18701},
  year      = {2025}
}

@article{matrix_game20,
  author    = {He, Xianglong and Peng, Chunli and Liu, Zexiang and others},
  title     = {{Matrix-Game} 2.0: An Open-source Real-Time and Streaming Interactive World Model},
  journal   = {arXiv preprint arXiv:2508.13009},
  year      = {2025}
}

@inproceedings{gamengen,
  author    = {Valevski, Dani and Leviathan, Yaniv and Arar, Moab and Fruchter, Shlomi},
  title     = {Diffusion Models Are Real-Time Game Engines},
  booktitle = ICLR,
  year      = {2025}
}

@article{yume15,
  author    = {Mao, Xiaofeng and Li, Zhen and Li, Chuanhao and Xu, Xiaojie and Ying, Kaining and He, Tong and Pang, Jiangmiao and Qiao, Yu and Zhang, Kaipeng},
  title     = {Yume-1.5: A Text-Controlled Interactive World Generation Model},
  journal   = {arXiv preprint arXiv:2512.22096},
  year      = {2025}
}

@article{hunyuan_world,
  author    = {Sun, Wenqiang and Zhang, Haiyu and Wang, Haoyuan and Wu, Junta and Wang, Zehan and Wang, Zhenwei and Wang, Yunhong and Zhang, Jun and Wang, Tengfei and Guo, Chunchao},
  title     = {{WorldPlay}: Towards Long-Term Geometric Consistency for Real-Time Interactive World Modeling},
  journal   = {arXiv preprint arXiv:2512.14614},
  year      = {2025}
}

@inproceedings{ddpm,
  author    = {Ho, Jonathan and Jain, Ajay and Abbeel, Pieter},
  booktitle = NeurIPS,
  title     = {Denoising Diffusion Probabilistic Models},
  volume    = {33},
  pages     = {6840--6851},
  year      = {2020}
}

@article{svd,
  title   = {Stable video diffusion: Scaling latent video diffusion models to large datasets},
  author  = {Blattmann, Andreas and Dockhorn, Tim and Kulal, Sumith and others},
  journal = {arXiv preprint arXiv:2311.15127},
  year    = {2023}
}

@inproceedings{ldm,
  author    = {Rombach, Robin and Blattmann, Andreas and Lorenz, Dominik and Esser, Patrick and Ommer, Bj\"orn},
  title     = {High-Resolution Image Synthesis With Latent Diffusion Models},
  booktitle = CVPR,
  year      = {2022},
  pages     = {10684--10695}
}

@article{3dgs,
  title   = {{3D Gaussian splatting} for real-time radiance field rendering},
  author  = {Kerbl, Bernhard and Kopanas, Georgios and Leimk{\"u}hler, Thomas and Drettakis, George},
  journal = TOG,
  volume  = {42},
  number  = {4},
  pages   = {139:1--139:14},
  year    = {2023}
}

@inproceedings{mesh_based_avatar_2,
  author    = {Alldieck, Thiemo and Magnor, Marcus and Bhatnagar, Bharat Lal and Theobalt, Christian and Pons-Moll, Gerard},
  title     = {Learning to Reconstruct People in Clothing From a Single {RGB} Camera},
  booktitle = CVPR,
  year      = {2019},
  pages     = {1175-1186}
}

@inproceedings{mesh_based_avatar_3,
  title     = {{SMPLX-Lite}: A realistic and drivable avatar benchmark with rich geometry and texture annotations},
  author    = {Jiang, Yujiao and Liao, Qingmin and Wang, Zhaolong and Lin, Xiangru and Lu, Zongqing and Zhao, Yuxi and Wei, Hanqing and Ye, Jingrui and Zhang, Yu and Shao, Zhijing},
  booktitle = ICME,
  pages     = {1--6},
  year      = {2024}
}

@inproceedings{nerf_avatar_1,
  title     = {{A-NeRF}: Articulated neural radiance fields for learning human shape, appearance, and pose},
  author    = {Su, Shih-Yang and Yu, Frank and Zollh{\"o}fer, Michael and Rhodin, Helge},
  booktitle = NeurIPS,
  volume    = {34},
  pages     = {12278--12291},
  year      = {2021}
}

@inproceedings{nerf_avatar_2,
  author    = {Guo, Chen and Jiang, Tianjian and Chen, Xu and Song, Jie and Hilliges, Otmar},
  title     = {{Vid2Avatar}: {3D} Avatar Reconstruction From Videos in the Wild via Self-Supervised Scene Decomposition},
  booktitle = CVPR,
  year      = {2023},
  pages     = {12858--12868}
}

@inproceedings{style_unet,
  author    = {Wang, Lizhen and Zhao, Xiaochen and Sun, Jingxiang and Zhang, Yuxiang and Zhang, Hongwen and Yu, Tao and Liu, Yebin},
  title     = {{StyleAvatar}: Real-time Photo-realistic Portrait Avatar from a Single Video},
  year      = {2023},
  booktitle = SIGGRAPH,
  pages     = {67:1--67:10}
}

@inproceedings{humanml3d,
  title     = {Generating diverse and natural {3D} human motions from text},
  author    = {Guo, Chuan and Zou, Shihao and Zuo, Xinxin and Wang, Sen and Ji, Wei and Li, Xingyu and Cheng, Li},
  booktitle = CVPR,
  pages     = {5152--5161},
  year      = {2022}
}

@inproceedings{amass,
  title     = {{AMASS}: Archive of motion capture as surface shapes},
  author    = {Mahmood, Naureen and Ghorbani, Nima and Troje, Nikolaus F and Pons-Moll, Gerard and Black, Michael J},
  booktitle = ICCV,
  pages     = {5442--5451},
  year      = {2019}
}

@article{kit_ml,
  title   = {The kit motion-language dataset},
  author  = {Plappert, Matthias and Mandery, Christian and Asfour, Tamim},
  journal = {Big Data},
  volume  = {4},
  number  = {4},
  pages   = {236--252},
  year    = {2016}
}

@inproceedings{pose_pred_1,
  title     = {On human motion prediction using recurrent neural networks},
  author    = {Martinez, Julieta and Black, Michael J and Romero, Javier},
  booktitle = CVPR,
  pages     = {2891--2900},
  year      = {2017}
}

@inproceedings{pose_pred_2,
  title     = {Human motion prediction via spatio-temporal inpainting},
  author    = {Hernandez, Alejandro and Gall, Jurgen and Moreno-Noguer, Francesc},
  booktitle = ICCV,
  pages     = {7134--7143},
  year      = {2019}
}

@inproceedings{keyframe_intepolation_1,
  title     = {Convolutional autoencoders for human motion infilling},
  author    = {Kaufmann, Manuel and Aksan, Emre and Song, Jie and Pece, Fabrizio and Ziegler, Remo and Hilliges, Otmar},
  booktitle = THREEDV,
  pages     = {918--927},
  year      = {2020}
}

@article{keyframe_intepolation_2,
  title   = {Robust motion in-betweening},
  author  = {Harvey, F{\'e}lix G and Yurick, Mike and Nowrouzezahrai, Derek and Pal, Christopher},
  journal = TOG,
  volume  = {39},
  number  = {4},
  pages   = {60:1--60:12},
  year    = {2020}
}

@inproceedings{mdm,
  title     = {Human Motion Diffusion Model},
  author    = {Tevet, Guy and Raab, Sigal and Gordon, Brian and Shafir, Yoni and Cohen-Or, Daniel and Bermano, Amit Haim},
  booktitle = ICLR,
  year      = {2023}
}

@inproceedings{emdm,
  title     = {{EMDM}: Efficient motion diffusion model for fast and high-quality motion generation},
  author    = {Zhou, Wenyang and Dou, Zhiyang and Cao, Zeyu and Liao, Zhouyingcheng and Wang, Jingbo and Wang, Wenjia and Liu, Yuan and Komura, Taku and Wang, Wenping and Liu, Lingjie},
  booktitle = ECCV,
  pages     = {18--38},
  year      = {2024}
}

@inproceedings{autokeyframe,
  title     = {{AutoKeyframe}: Autoregressive keyframe generation for human motion synthesis and editing},
  author    = {Zheng, Bowen and Chen, Ke and Yao, Yuxin and Zeng, Zijiao and Jiang, Xinwei and Wang, He and Lasenby, Joan and Jin, Xiaogang},
  booktitle = SIGGRAPH,
  pages     = {12:1--12:12},
  year      = {2025}
}

@inproceedings{vista,
  title={Vista: A generalizable driving world model with high fidelity and versatile controllability},
  author={Gao, Shenyuan and Yang, Jiazhi and Chen, Li and Chitta, Kashyap and Qiu, Yihang and Geiger, Andreas and Zhang, Jun and Li, Hongyang},
  booktitle=NeurIPS,
  volume={37},
  pages={91560--91596},
  year={2024}
}

@inproceedings{genie,
  title={Genie: Generative interactive environments},
  author={Bruce, Jake and Dennis, Michael and Edwards, Ashley and others},
  booktitle=ICML,
  pages={4603--4623},
  year={2024}
}

@article{wham,
  title={World and human action models towards gameplay ideation},
  author={Kanervisto, Anssi and Bignell, Dave and Wen, Linda Yilin and others},
  journal=Nature,
  volume={638},
  number={8051},
  pages={656--663},
  year={2025}
}

@misc{genie2,
  title={Genie 2: A large-scale foundation world model},
  author={Jack Parker-Holder and Philip Ball and Jake Bruce and others
  },
  year={2024},
  howpublished={\url{https://deepmind.google/discover/blog/genie-2-a-large-scale-foundation-world-model/}}
}

@misc{genie3,
  title={Genie 3: A new frontier for world models},
  author={Philip J. Ball and Jakob Bauer and Frank Belletti and others},
  year={2025},
  howpublished={\url{https://deepmind.google/blog/genie-3-a-new-frontier-for-world-models/}}
}

@inproceedings{sds,
  title={{DreamFusion}: Text-to-{3D} using {2D} Diffusion},
  author={Poole, Ben and Jain, Ajay and Barron, Jonathan T. and Mildenhall, Ben},
  booktitle=ICLR,
  year={2023},
}

@inproceedings{sd30,
  title={Scaling Rectified Flow Transformers for High-Resolution Image Synthesis},
  author={Esser, Patrick and Kulal, Sumith and Blattmann, Andreas and others},
  booktitle=ICML,
  year={2024},
  pages={12606--12633},
}

@inproceedings{imagen,
  title={Photorealistic text-to-image diffusion models with deep language understanding},
  author={Saharia, Chitwan and Chan, William and Saxena, Saurabh and others},
  booktitle=NeurIPS,
  volume={35},
  pages={36479--36494},
  year={2022}
}

@inproceedings{sv3d,
  title={{SV3D}: Novel multi-view synthesis and {3D} generation from a single image using latent video diffusion},
  author={Voleti, Vikram and Yao, Chun-Han and Boss, Mark and Letts, Adam and Pankratz, David and Tochilkin, Dmitry and Laforte, Christian and Rombach, Robin and Jampani, Varun},
  booktitle=ECCV,
  pages={439--457},
  year={2024}
}

@article{videofactory,
  title={{VideoFactory}: Swap attention in spatiotemporal diffusions for text-to-video generation},
  author={Wang, Wenjing and Yang, Huan and Tuo, Zixi and He, Huiguo and Zhu, Junchen and Fu, Jianlong and Liu, Jiaying},
  journal={arXiv preprint arXiv:2305.10874},
  year={2023}
}

@inproceedings{mvdream,
  title={{MVDream}: Multi-view Diffusion for {3D} Generation},
  author={Shi, Yichun and Wang, Peng and Ye, Jianglong and Mai, Long and Li, Kejie and Yang, Xiao},
  booktitle=ICLR,
  year={2024},
}

@inproceedings{sds4d_1,
  title={Text-to-{4D} dynamic scene generation},
  author={Singer, Uriel and Sheynin, Shelly and Polyak, Adam and others},
  booktitle=ICML,
  pages={31915--31929},
  year={2023}
}

@inproceedings{sds4d_2,
  title={{Consistent4D}: Consistent 360{\textdegree} Dynamic Object Generation from Monocular Video},
  author={Jiang, Yanqin and Zhang, Li and Gao, Jin and Hu, Weiming and Yao, Yao},
  booktitle=ICLR,
  year={2024},
}

@inproceedings{diffusion2,
  title={Diffusion\textsuperscript{2}: Dynamic {3D} Content Generation via Score Composition of Video and Multi-view Diffusion Models},
  author={Yang, Zeyu and Pan, Zijie and Gu, Chun and Zhang, Li},
  booktitle=ICLR,
  year={2025},
}

@inproceedings{dnerf,
  title={{D-NeRF}: Neural radiance fields for dynamic scenes},
  author={Pumarola, Albert and Corona, Enric and Pons-Moll, Gerard and Moreno-Noguer, Francesc},
  booktitle=CVPR,
  pages={10318--10327},
  year={2021}
}

@article{4dgs,
  title={{WideRange4D}: Enabling High-Quality {4D} Reconstruction with Wide-Range Movements and Scenes},
  author={Yang, Ling and Zhu, Kaixin and Tian, Juanxi and Zeng, Bohan and Lin, Mingbao and Pei, Hongjuan and Zhang, Wentao and Yan, Shuichen},
  journal={arXiv preprint arXiv:2503.13435},
  year={2025}
}

@inproceedings{shapegen4d,
  title={{ShapeGen4D}: Towards High Quality {4D} Shape Generation from Videos},
  author={Yenphraphai, Jiraphon and Mirzaei, Ashkan and Chen, Jianqi and Zou, Jiaxu and Tulyakov, Sergey and Yeh, Raymond A. and Wonka, Peter and Wang, Chaoyang},
  booktitle=ICLR,
  year={2026}
}

@inproceedings{ptv3,
  title={{Point Transformer V3}: Simpler faster stronger},
  author={Wu, Xiaoyang and Jiang, Li and Wang, Peng-Shuai and Liu, Zhijian and Liu, Xihui and Qiao, Yu and Ouyang, Wanli and He, Tong and Zhao, Hengshuang},
  booktitle=CVPR,
  pages={4840--4851},
  year={2024}
}

@inproceedings{splatformer,
    title={SplatFormer: Point Transformer for Robust {3D} Gaussian Splatting},
    author={Yutong Chen and Marko Mihajlovic and Xiyi Chen and Yiming Wang and Sergey Prokudin and Siyu Tang},
    booktitle=ICLR,
    year={2025},
}

@inproceedings{unet,
  title={{U-Net}: Convolutional networks for biomedical image segmentation},
  author={Ronneberger, Olaf and Fischer, Philipp and Brox, Thomas},
  booktitle=MICCAI,
  pages={234--241},
  year={2015}
}

@inproceedings{cfg,
  title={Classifier-Free Diffusion Guidance},
  author={Ho, Jonathan and Salimans, Tim},
  booktitle=NeurIPSW,
  year={2021},
}

@inproceedings{diffusion_forcing,
  title={{Diffusion Forcing}: Next-token prediction meets full-sequence diffusion},
  author={Chen, Boyuan and Mart{\'\i} Mons{\'o}, Diego and Du, Yilun and Simchowitz, Max and Tedrake, Russ and Sitzmann, Vincent},
  booktitle=NeurIPS,
  volume={37},
  pages={24081--24125},
  year={2024}
}

@inproceedings{ddim,
  title={Denoising Diffusion Implicit Models},
  author={Song, Jiaming and Meng, Chenlin and Ermon, Stefano},
  booktitle=ICLR,
  year={2021},
}

@article{fvd,
  title={Towards accurate generative models of video: A new metric \& challenges},
  author={Unterthiner, Thomas and Van Steenkiste, Sjoerd and Kurach, Karol and Marinier, Raphael and Michalski, Marcin and Gelly, Sylvain},
  journal={arXiv preprint arXiv:1812.01717},
  year={2018}
}

@misc{mixamo,
  author={Adobe},
  title={Mixamo},
  howpublished={\url{https://www.mixamo.com/}},
  year={2026},
  note={Accessed: 2026-01-21}
}

@InProceedings{i3d,
    author = {Carreira, Joao and Zisserman, Andrew},
    title = {Quo Vadis, Action Recognition? A New Model and the Kinetics Dataset},
    booktitle = CVPR,
    year = {2017}
}

@inproceedings{vqvae,
 title={Neural discrete representation learning},
 author = {van den Oord, Aaron and Vinyals, Oriol and kavukcuoglu, koray},
 booktitle = NeurIPS,
 pages = {6309--6318},
 volume = {30},
 year = {2017}
}

@article{st_transformer,
  title={Spatial-temporal transformer networks for traffic flow forecasting},
  author={Xu, Mingxing and Dai, Wenrui and Liu, Chunmiao and Gao, Xing and Lin, Weiyao and Qi, Guo-Jun and Xiong, Hongkai},
  journal={arXiv preprint arXiv:2001.02908},
  year={2020}
}

@inproceedings{vae,
  title={Auto-encoding variational bayes},
  author={Kingma, Diederik P and Welling, Max},
  booktitle=ICLR,
  year={2013}
}

@article{ssim,
  title={Image quality assessment: from error visibility to structural similarity},
  author={Wang, Zhou and Bovik, Alan C and Sheikh, Hamid R and Simoncelli, Eero P},
  journal=TIP,
  volume={13},
  number={4},
  pages={600--612},
  year={2004},
}

@inproceedings{lpips,
  title={The unreasonable effectiveness of deep features as a perceptual metric},
  author={Zhang, Richard and Isola, Phillip and Efros, Alexei A and Shechtman, Eli and Wang, Oliver},
  booktitle=CVPR,
  pages={586--595},
  year={2018}
}

\newpage
\setcounter{section}{0}
\setcounter{figure}{0}
\setcounter{table}{0}
\setcounter{equation}{0}
\renewcommand{\thesection}{\Alph{section}}
\renewcommand{\thesubsection}{\Alph{section}.\arabic{subsection}}
\renewcommand{\thesubsubsection}{\Alph{section}.\arabic{subsection}.\arabic{subsubsection}}
\renewcommand{\thefigure}{\Alph{figure}}
\renewcommand{\thetable}{\Alph{table}}

\section*{Supplementary Material}
\section{Dataset Details}

\begin{figure*}[t]
	\centering
	\includegraphics[width=\textwidth]{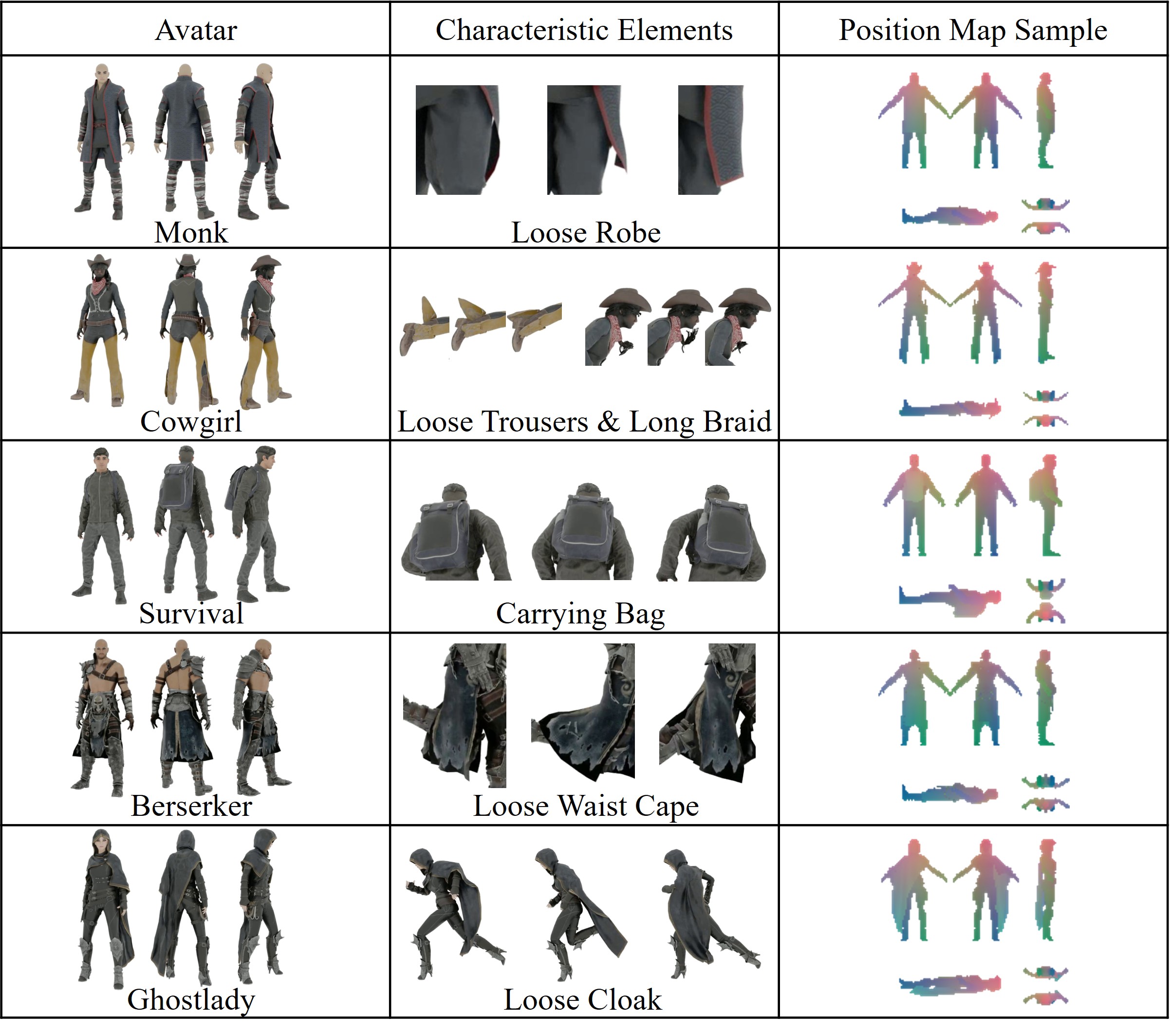}
    \vspace{-0.15in}
    \caption{Overview of the five avatars exported from UE. Each avatar exhibits characteristic non-rigid or non-body elements (\eg, loose garments, braids, and carried accessories) that go beyond a standard body mesh.}
    \vspace{-0.15in}
	\label{fig:dataset}
\end{figure*}

We export mesh sequences of five distinct avatars from Unreal Engine (UE), namely \textit{Monk}, \textit{Cowgirl}, \textit{Survival}, \textit{Berserker}, and \textit{Ghostlady}. An overview of these avatars and their characteristic elements is provided in \cref{fig:dataset}.

All avatars are retargeted to the locomotion system of the \textit{GameAnimationSample} project, from which we export motion sequences under five action signals: Forward~\key{W}, Left~\key{A}, Backward~\key{S}, Right~\key{D}, and Idle~\key{$\emptyset$}. Specifically, we capture all pairwise transitions between different key presses (\eg, Forward to Backward, \key{W}$\to$\key{S}). The number of frames for each transition pair is reported in \cref{tab:sequence_length}, yielding a total of 6{,}253 frames across all sequences. Within each sequence, the avatar begins in steady-state locomotion under the source action signal, switches to the target signal at a designated turning point, and then continues moving under the new key press for a sustained period. The sequences are sufficiently long that, apart from a brief transient phase immediately following the turning point, the remaining frames capture the avatar in stable directional locomotion. Consequently, although constant-key sequences (\eg, holding \key{W} throughout) are not explicitly exported as separate clips, they are implicitly contained within the recorded sequences.

\begin{table}[ht]
\centering
\caption{Number of frames in each transition sequence from one key press to another. Each entry corresponds to a sequence in which the avatar transitions from the start key (column) to the end key (row).}
\label{tab:sequence_length}
\begin{tabularx}{0.7\textwidth}{c|YYYYY}
\hline
\diagbox{End Key}{Start Key} & \key{W} & \key{A} & \key{S} & \key{D} & \key{$\emptyset$} \\
\hline
\key{W}          & --  & 304 & 330 & 308 & 264 \\
\key{A}          & 312 & --  & 307 & 329 & 234 \\
\key{S}          & 334 & 306 & --  & 304 & 264 \\
\key{D}          & 310 & 328 & 309 & --  & 260 \\
\key{$\emptyset$} & 367 & 359 & 362 & 362 & --  \\
\hline
\end{tabularx}
\end{table}

\section{Additional Method Details}
\subsection{Coarse-to-Fine Appearance Mapping}
\subsubsection{Base Attribute Map.}
For each avatar, we first train its base 3D Gaussian Splatting (3DGS)~\cite{3dgs} attribute map on a textured mesh in a standing pose. The mesh is rendered from 50 viewpoints distributed along a cylindrical surface. For rendering, the mesh is normalized to an Axis-Aligned Bounding Box (AABB) of \([-0.5, 0.5]\), which differs from the transformations used in group-wise normalization. Accordingly, for the position map corresponding to the standing pose, we first upscale it to \(1024 \times 1024\) and then normalize the recovered point cloud using the same AABB. The AABB-normalized points serve as the means of the Gaussian ellipsoids, which remain frozen during base attribute optimization. All other attributes are initialized following the standard 3DGS procedure and optimized using a rendering loss:
\begin{equation}
	\mathcal{L}_{\text{render}} = \lambda_1 \|\mathbf{I}_{\text{gt}} - \mathbf{I}_{\text{r}}\|_1 + \lambda_2 \left(1 - \text{SSIM}(\mathbf{I}_{\text{gt}}, \mathbf{I}_{\text{r}})\right) + \lambda_3 \mathcal{L}_{\text{LPIPS}}(\mathbf{I}_{\text{gt}}, \mathbf{I}_{\text{r}}),
	\label{eq:render_loss}
\end{equation}
where \(\mathbf{I}_{\text{gt}}\) and \(\mathbf{I}_{\text{r}}\) denote the ground-truth and rendered images, respectively, \(\text{SSIM}(\cdot, \cdot)\)~\cite{ssim} computes the structural similarity index, \(\mathcal{L}_{\text{LPIPS}}\) ~\cite{lpips} measures perceptual similarity, and \(\lambda_1\), \(\lambda_2\), \(\lambda_3\) are weighting coefficients. Densification is disabled during base attribute optimization to maintain a consistent point count, and we use rank-zero Spherical Harmonics (SH). The resulting base attributes fit the standing pose accurately and provide a strong initialization.

\subsubsection{PTv3.}
The modified PTv3 model~\cite{ptv3, splatformer} is trained across all poses in the training set to learn pose-dependent attribute refinements. For a given pose, analogous to the standing-pose procedure, we load its six-view position map, convert it to a point cloud, and normalize it with the \([-0.5, 0.5]\) AABB. The normalized points are then shifted to $[0, 1]$ and used as means of the coarse 3DGS \(\mathcal{G}_{\text{coarse}}\). Because the longest axis (avatar height) shortens when the avatar bends, the same avatar is scaled differently across poses during normalization. Let \(\mathcal{H}_s\) denote the height of the avatar in the standing pose used for base Gaussian attribute training. To scale the Gaussian ellipsoids proportionally with the avatar, we multiply the base scaling attribute by \(\mathcal{H}_s / \mathcal{H}_p\) for a posed avatar with original height \(\mathcal{H}_p\). The remaining attributes, \ie, SH, rotation, and opacity, are directly assigned to \(\mathcal{G}_{\text{coarse}}\).

In the PTv3 model, the 14-channel Gaussian attributes (3 for means, 3 for SH, 3 for scaling, 4 for rotation, and 1 for opacity) are stacked and projected into embeddings via a Multi-Layer Perceptron (MLP). Taking these embeddings as input features, \(\mathcal{G}_{\text{coarse}}\) is passed through the PTv3 backbone. Separate prediction heads for means, SH, scaling, rotation, and opacity predict per-attribute offsets. The refined 3DGS \(\mathcal{G}_{\text{refined}}\) is then rendered for PTv3 training with the same rendering loss as in \cref{eq:render_loss}.

\subsection{PCA-Based Position Map Alignment}
\subsubsection{Formulation.}

As shown in the ablation study in the main paper, applying Principal Component Analysis (PCA) to align the six rendered views in generated position maps leads to noticeable improvements in the visual quality of \methodName{}. In this section, we provide the technical details of this process and discuss why it is effective. While~\cite{animatable_gaussians} uses the same PCA procedure on position maps, their intention is to project novel poses into the training distribution. On the other hand, our objective is to align the six views of the position maps.

From each four-frame group, we extract the fourth position map as a training sample for PCA. For each position map, all foreground pixels corresponding to valid 3D coordinates are gathered and flattened into a single vector \(\mathbf{x}_j \in \mathbb{R}^{3N}\), where \(N\) is the number of foreground pixels. Stacking all \(F\) training vectors yields the data matrix \(\mathbf{X} = [\mathbf{x}_1, \dots, \mathbf{x}_F]\) and the sample mean:
\begin{equation}
    \bar{\mathbf{x}}
    = \frac{1}{F} \sum_{j=1}^{F} \mathbf{x}_j
\end{equation}
We perform PCA on \(\mathbf{X}\) and retain the top \(M\) principal components as columns of the basis matrix:
\begin{equation}
\label{eq:pca_components}
    \mathbf{B}
    = [\mathbf{b}_1, \mathbf{b}_2, \dots, \mathbf{b}_M]
    \in \mathbb{R}^{3N \times M}
\end{equation}

During inference, the foreground pixels of a generated position map are flattened into \(\mathbf{x}_{\text{gen}}\) in the same manner. We first project it onto the learned basis to obtain the coefficient vector:
\begin{equation}
\label{eq:pca_coeff}
    \mathbf{w}
    = \mathbf{B}^T (\mathbf{x}_{\text{gen}} - \bar{\mathbf{x}})
    \in \mathbb{R}^{M}
\end{equation}
The aligned position map is then reconstructed as:
\begin{equation}
\label{eq:pca_recon}
    \mathbf{x}_{\text{recon}}
    = \mathbf{B}\,\mathbf{w} + \bar{\mathbf{x}}
\end{equation}

Because the principal components \(\mathbf{B}\) and the mean \(\bar{\mathbf{x}}\) are derived entirely from ground-truth position maps rendered from meshes, the subspace spanned by \(\mathbf{B}\) captures only modes of variation that preserve six-view consistency. In \cref{eq:pca_coeff}, the projection onto \(\mathbf{B}\) retains only the portion of \(\mathbf{x}_{\text{gen}}\) that lies within this consistent subspace, while any component falling outside it---including cross-view misalignment introduced during generation---is discarded. Consequently, \(\mathbf{x}_{\text{recon}}\) in \cref{eq:pca_recon} represents the closest six-view-consistent position map to \(\mathbf{x}_{\text{gen}}\) in the least-squares sense.

\subsubsection{Analysis of Principal Components.}
While PCA reconstruction helps align the six views of the position maps, it is important that this alignment does not observably alter the avatar pose, as doing so would render the motion sequence unnatural. Therefore, we analyze how many principal components are needed for sufficient reconstruction accuracy. 

\begin{figure*}[t]
	\centering
	\includegraphics[width=0.8\textwidth]{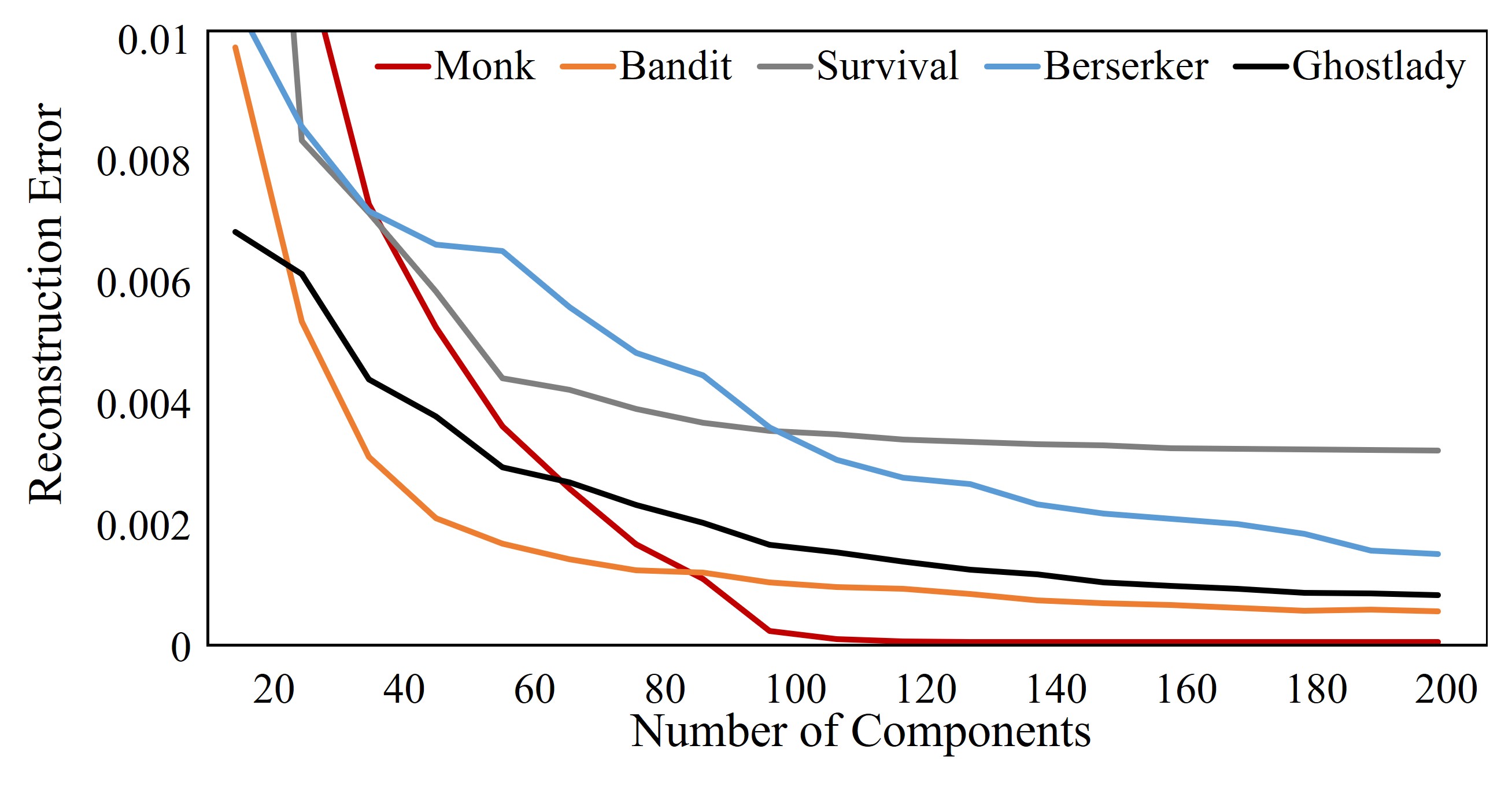}
    \vspace{-0.1in}
    \caption{Reconstruction error as a function of the number of principal components.}
	\label{fig:pca_components}
\end{figure*}

For each avatar, we use a generated position map as the reconstruction target and measure the reconstruction error as a function of the number of principal components in \cref{fig:pca_components}. The line chart shows that the rate of decrease in reconstruction error for each avatar levels off at approximately 120 principal components. However, the reconstruction error continues to gradually decrease beyond that point. In practice, we use 200 components to ensure sufficient accuracy. As reported in \cref{tab:runtime}, the computational overhead of PCA reconstruction remains negligible despite the large number of components.

\begin{figure*}[t]
	\centering
	\includegraphics[width=\textwidth]{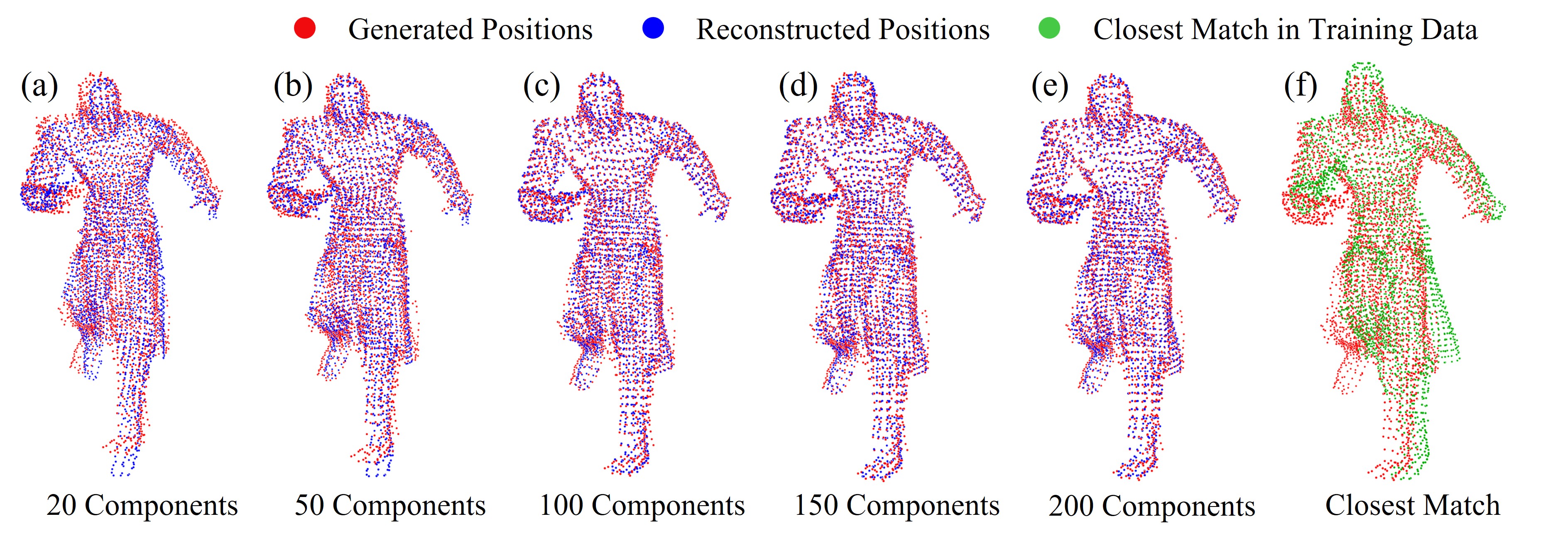}
    \vspace{-0.15in}
    \caption{Visualization of PCA reconstruction with varying numbers of principal components.}
    \vspace{-0.15in}
	\label{fig:pca_vis}
\end{figure*}

We visualize the point clouds recovered from generated and reconstructed position maps with varying numbers of principal components in \cref{fig:pca_vis}(a)--(e). As shown, when using 20 components as in~\cite{animatable_gaussians}, there remains a non-negligible discrepancy between the generated geometry and its reconstruction. As the number of principal components increases, this discrepancy diminishes progressively, until the PCA reconstruction yields a geometry that closely approximates the generated pose while eliminating inaccurate positions. 

In \cref{fig:pca_vis}(f), we visualize the avatar with the closest geometry found in the training data, showing that the generated avatar assumes a novel pose that differs noticeably from this closest match. This confirms that \methodName{} does not simply replicate training poses but rather generates novel ones conditioned on temporal context and the action signal in a physically plausible manner. It also demonstrates that PCA reconstruction preserves the generated poses rather than collapsing them toward the training distribution.

\section{Implementation Details}

\subsection{Network Architecture.} 

\subsubsection{VAE.}
The denoising UNet in \methodName{} operates in the latent space produced by a Variational AutoEncoder (VAE)~\cite{vae}, which encodes 3-channel position maps into 4-channel latent representations with a spatial downsampling factor of 8. The VAE encoder and decoder each comprise four blocks with channel dimensions $[128, 256, 512, 512]$ and two residual layers per block. Three out of the four encoder blocks apply downsampling, mirroring the three upsampling blocks in the decoder.

\subsubsection{UNet.}
The UNet takes 4-channel latent inputs and processes them through four encoder stages and four symmetric decoder stages with skip connections, with block channel dimensions $[320, 640, 1280, 1280]$. Each encoder stage contains two residual blocks, while each decoder stage contains three. In the first three encoder stages and the last three decoder stages, every residual block is paired with a transformer block that performs spatial attention, cross attention, and temporal attention. The mid block inserts one such transformer block between two residual blocks. The first three encoder stages each apply $2\times$ spatial downsampling, with the spatial resolution correspondingly restored in the decoder. Action conditioning is injected through cross attention layers with 768-dimensional action embeddings.

\subsubsection{PTv3.}
The modified PTv3 model first discretizes Gaussian positions onto a $192^3$ voxel grid for serialization. The 14-channel input attributes are projected into 32-dimensional embeddings via an MLP. The encoder comprises five stages with channel dimensions $[32, 64, 128, 256, 512]$, depths $[2, 2, 2, 6, 2]$, and serialized pooling strides of $[1, 2, 2, 2]$ between consecutive stages. The symmetric decoder has four stages with channel dimensions $[64, 64, 128, 256]$, each of depth 2, and uses skip connections from the corresponding encoder stages. All attention blocks use four serialization orders (z, z-trans, Hilbert, Hilbert-trans) with a patch size of 1024. The 64-dimensional decoder output is concatenated with the 14-dimensional input attributes and fed to per-attribute prediction heads, each a 4-layer MLP with a hidden width of 128. The final linear layer of each head is zero-initialized so that the model starts as an identity mapping with no 3DGS refinement. A tanh activation is applied to the predicted means offset to bound spatial corrections, while all other attribute offsets are unconstrained.

\subsection{Training Configuration}
We provide the training configurations for all models in \methodName{}, which are trained on 8 NVIDIA A6000 GPUs. The VAE is trained for 300k iterations with a batch size of 32 and a learning rate of 1.5e-5. The denoising UNet is trained for 100k and 200k iterations in the first and second stages, respectively, totaling 300k iterations with a batch size of 32 and a learning rate of 5e-5. Twenty-five frames immediately following each turning point are oversampled in the second stage with a probability of 70\%, while the remaining 30\% is reserved for non-turning frames. PTv3 is trained for 20 epochs with a batch size of 8 and a learning rate of 1e-4. In each iteration, 8 out of 50 views are randomly selected for supervision. We observe that consistently using a black background for the rendering loss causes black Gaussian ellipsoids to protrude from the body surface. To address this, we randomly alternate between black and white backgrounds at each iteration, which effectively eliminates the artifact.

\subsection{Baseline Implementations}

\subsubsection{Animatable Gaussians.} Animatable Gaussians~\cite{animatable_gaussians} requires multi-view videos and per-frame SMPL-X\cite{smplx} parameters as input for both training and inference. To this end, we fit SMPL-X parameters to the motion sequences of the five avatars. Specifically, we render the meshes into multi-view videos and use EasyMocap~\cite{easymocap} for SMPL-X optimization. Training of Animatable Gaussians follows the official open-source implementation.

\subsubsection{Mixamo.} For Mixamo~\cite{mixamo}, we manually upload the A-pose mesh of each avatar to their website and export the rigging results. The exported rigged meshes are then driven using the ground-truth skeleton sequences. Consequently, the Mixamo results share identical skeleton sequences with the ground truth but differ in skinning weights.

\subsubsection{SV4D 2.0.} We provide SV4D 2.0~\cite{sv4d20} with front-view videos rendered from the ground-truth meshes as input for 4D generation. SV4D 2.0 is configured to generate 21-frame videos from four cardinal viewpoints using its pretrained checkpoint.

\subsection{FVD Measurement}
\begin{sloppypar}
When measuring Fréchet Video Distance (FVD)\cite{fvd}, all baselines and our method are compared against the same ground-truth videos, which are rendered mesh sequences from four cardinal viewpoints. The primary purpose of FVD measurement is to evaluate the autoregressive performance of \methodName{}. As reported in the main paper, we autoregressively generate a 200-frame trajectory \{60$\times$\key{W}, 60$\times$\key{S}, 60$\times$\key{A}, 20$\times$\key{$\emptyset$}\} for \methodName{}, encompassing one start, one $180^\circ$ turn, one $90^\circ$ turn, and one stop. For each avatar, the sequence starts from a position map corresponding to a standing pose, which is repeated three times and used as the initial input of \methodName{}. Given that the ground-truth sequence under equivalent action signals is not exported (\cref{tab:sequence_length}), we clip a total of 200 frames from sequences \key{$\emptyset$}$\to$\key{W} (start), \key{W}$\to$\key{S} ($180^\circ$ turn), \key{W}$\to$\key{A} ($90^\circ$ turn), and \key{W}$\to$\key{$\emptyset$} (stop) to form the ground-truth set of videos. Estimated SMPL-X parameters and ground-truth skeletal poses of these clipped frames are provided to Animatable Gaussians\cite{animatable_gaussians} and Mixamo\cite{mixamo}, respectively, to generate their prediction sets of videos. During FVD calculation, video features are extracted with the Inflated 3D ConvNet~\cite{i3d} model with non-overlapping 16-frame windows and batch size 8. The feature extraction setting is consistent for ground-truth videos, \methodName{}, and the baselines.
\end{sloppypar}

\section{Additional Results and Analysis}

\subsection{Ablation Studies}

\begin{figure*}[t]
	\centering
	\includegraphics[width=\textwidth]{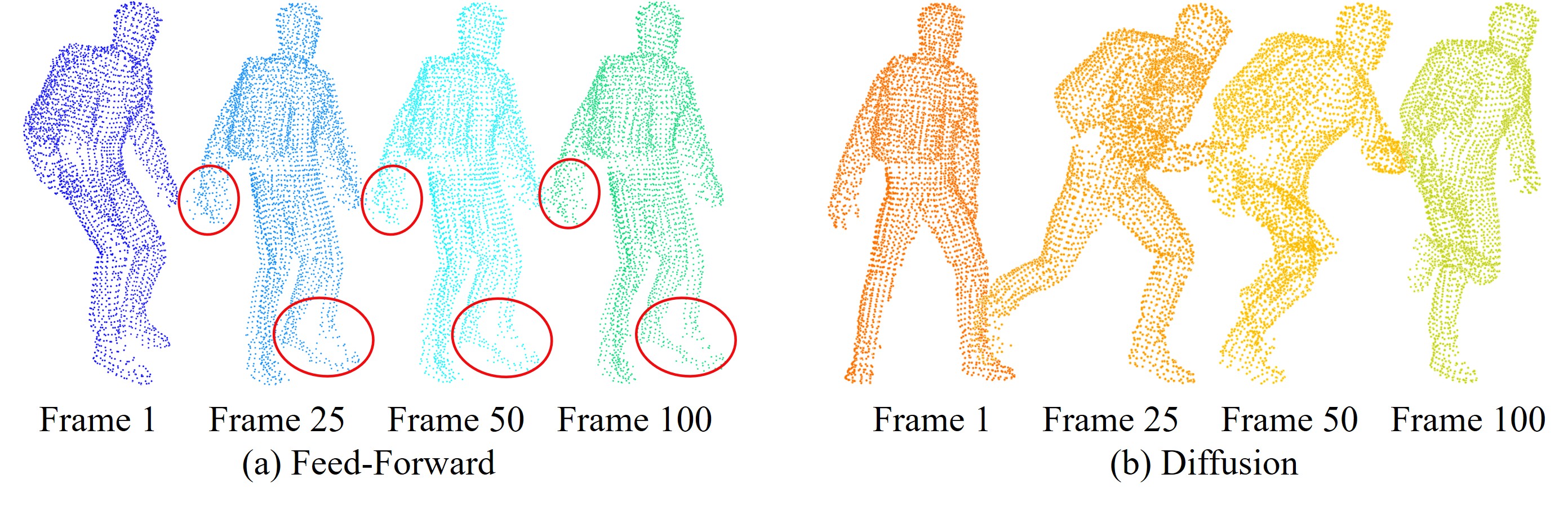}
    \vspace{-0.2in}
    \caption{Comparison between the feed-forward pipeline and the diffusion pipeline. PCA reconstruction is disabled to show raw model outputs.}
	\label{fig:paradigm_comp}
\end{figure*}

\subsubsection{Feed-Forward vs.\ Diffusion.} An intuitive option for the action-driven neural avatar task is to adopt a feed-forward pipeline, where a feed-forward model predicts the next state of the avatar based on the action signal. To compare the feed-forward pipeline with the proposed multi-frame diffusion pipeline, we experiment with a combination of a Vector Quantized Variational AutoEncoder (VQ-VAE)\cite{vqvae} and a Spatial-Temporal Transformer (ST-Transformer)\cite{st_transformer} following Genie~\cite{genie}. Specifically, the first three frames of position maps are encoded into the codebook space by the VQ-VAE. The fourth frame is zero-initialized, concatenated with the first three frames, and passed to the ST-Transformer. The action embedding is added to the tokens of all four input frames, and the ST-Transformer progressively updates the tokens of the fourth frame based on temporal context and the action embedding. The updated tokens of the fourth frame are then aligned to the codebook space and decoded into a position map. Results are shown in \cref{fig:paradigm_comp}, where we deliberately disable PCA reconstruction to expose the raw outputs of both models. As illustrated in the figure, the feed-forward pipeline fails in two respects. First, the geometry quality is severely degraded: points on the hands and legs are largely misaligned. Second, the pipeline fails to drive the avatar, which mostly remains stationary. Based on these observations, we adopt the multi-frame diffusion paradigm in \methodName{}.

\begin{figure*}[t]
	\centering
	\includegraphics[width=\textwidth]{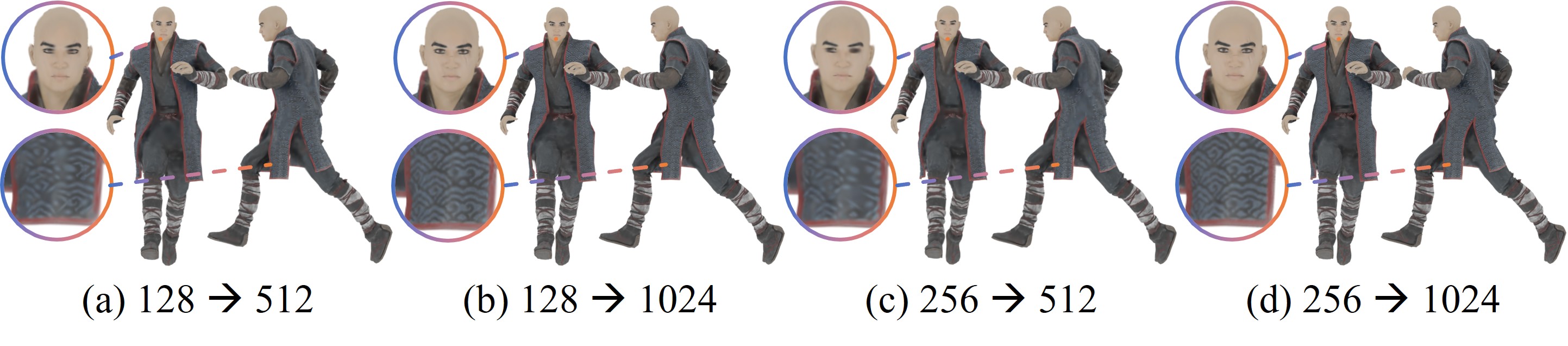}
    \vspace{-0.2in}
    \caption{Comparison of avatar appearance trained under different resolution configurations. (a) $128\times128$ position maps upscaled to $512\times512$. (b) $128\times128$ position maps upscaled to $1024\times1024$. (c) $256\times256$ position maps upscaled to $512\times512$. (d) $256\times256$ position maps upscaled to $1024\times1024$. All upscaling is performed via foreground-aware bilinear interpolation.}
    \vspace{-0.15in}
	\label{fig:res_comp}
\end{figure*}

\subsubsection{Position Map Resolution.} Recall that \methodName{} adopts a low-resolution geometry and high-resolution appearance paradigm. We show the results of avatar appearance trained under four resolution configurations in \cref{fig:res_comp}. As shown in the figure, avatar appearance depends heavily on the position map resolution (\ie, point count) after bilinear interpolation. By contrast, the position map resolution before upscaling has little impact on the final results. Although the point cloud recovered from the position map upsampled from $128\times128$ resolution exhibits less geometric smoothness, PTv3 can predict offsets for the Gaussian means, reducing aliasing artifacts and enabling 3DGS to fit the avatar appearance with high fidelity. We thereby demonstrate the feasibility of synthesizing high-fidelity avatar appearance from position maps at a resolution as low as $128\times128$. On the other hand, training a latent diffusion model on $256\times256$ position maps is considerably more expensive. Our low-resolution geometry and high-resolution appearance design therefore achieves a favorable balance between quality and computational cost.

\subsubsection{Number of DDIM Sampling Steps.} We report the performance of \methodName{} with respect to the number of DDIM~\cite{ddim} steps in \cref{tab:ddim}. While DDIM sampling is applied exclusively in the geometry stage of \methodName{}, varying the number of DDIM steps also influences the visual quality produced by the same PTv3 during appearance mapping. As shown in \cref{tab:ddim}, the visual quality of \methodName{} gradually improves with an increasing number of DDIM steps but plateaus at 10 steps. This finding is consistent with the observation in~\cite{gamengen} that, when the data space of the diffusion model is restricted to position maps of the same avatar and the generation process is heavily conditioned on both temporal context and the action signal, fewer sampling steps suffice. These characteristics allow \methodName{} to enjoy reduced inference time and make it amenable to further acceleration techniques.

\begin{table}[t]
  \centering
  \caption{Model performance with respect to the number of DDIM steps. Best results are highlighted in \textbf{bold}.}
  \vspace{-0.05in}
    \begin{tabularx}{0.8\textwidth}{lYYYYY}
      \toprule
       & 1 step & 2 steps & 4 steps & 10 steps & 20 steps \\
      \midrule
      FVD$\downarrow$   & 402.88 & 404.58 & 248.32 & \textbf{194.87} & 196.05 \\
      PSNR$\uparrow$    & 19.26  & 23.02  & 23.48  & \textbf{23.95}  & 23.90  \\
      LPIPS$\downarrow$  & 0.171  & 0.118  & 0.113  & \textbf{0.106}  & 0.109  \\
      \bottomrule
    \end{tabularx}%
  \label{tab:ddim}
\end{table}

\subsection{Per-Avatar Comparisons}

\begin{figure*}[t]
	\centering
	\includegraphics[width=\textwidth]{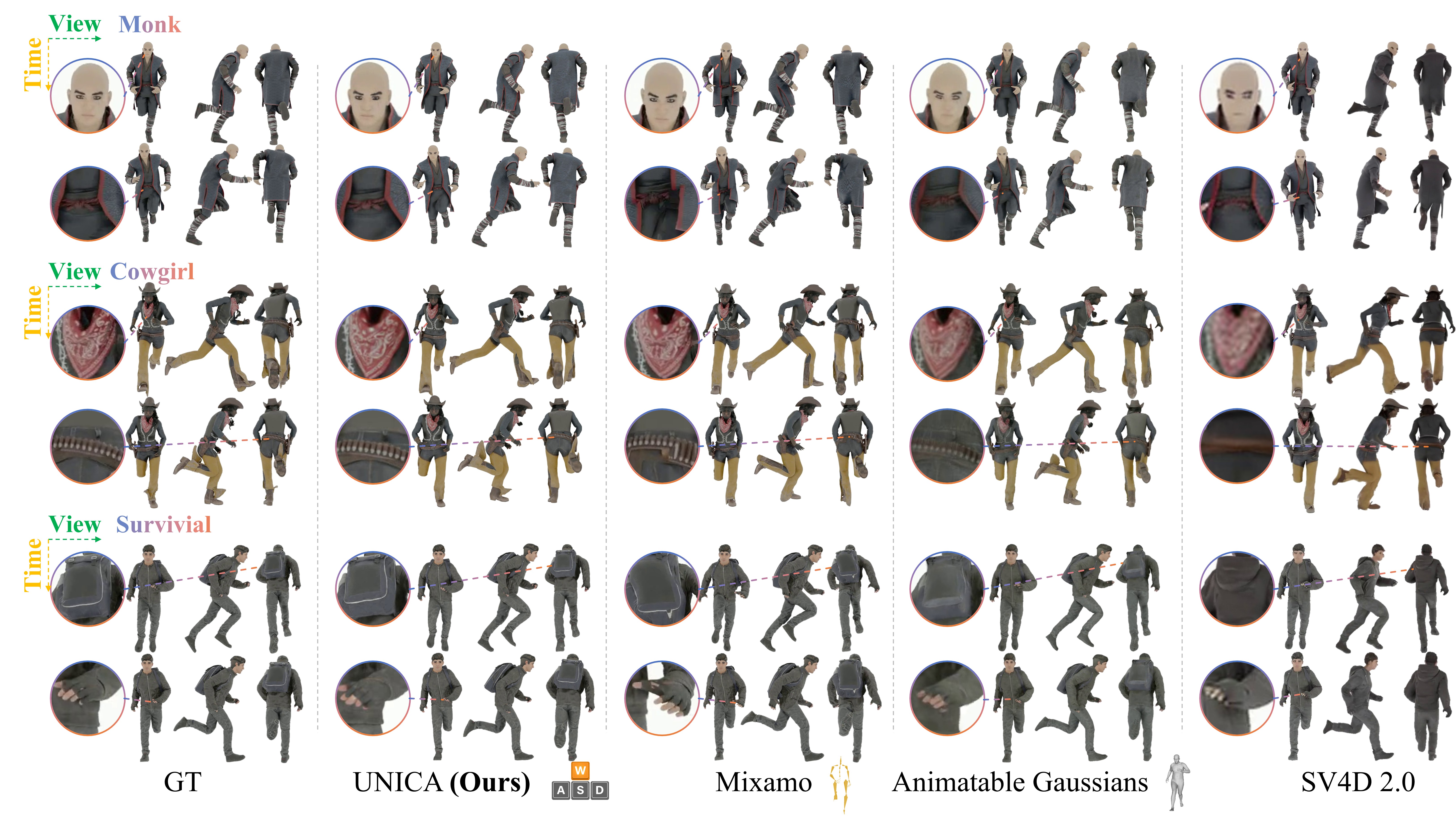}
    \vspace{-0.15in}
    \caption{Additional qualitative comparisons between \methodName{} and baseline methods across different frames and viewpoints.}
	\label{fig:supp_vis_comp}
\end{figure*}

\begin{table}[tb]
  \caption{Quantitative comparison across five avatars. Best results are highlighted in \textbf{bold}. PSNR and LPIPS are reported as mean $\pm$ standard deviation.}
  \vspace{-0.05in}
  \label{tab:supp_comparison}
  \centering
  \begin{tabular}{@{\hspace{6pt}}ll*{3}{>{\centering\arraybackslash}p{2.8cm}}@{}}
    \toprule
    Case & Metric & Ani.\ Gaussians~\cite{animatable_gaussians} & Mixamo~\cite{mixamo} & \methodName{} (Ours) \\
    \midrule
    \multirow{3}{*}{Monk}   & FVD $\downarrow$ & \textbf{136.83} & 145.97 & 156.92\\
              & PSNR $\uparrow$  & 22.45 $\pm$3.23 & 16.63 $\pm$1.27 & \textbf{23.21 $\pm$2.16}\\
              & LPIPS $\downarrow$ & \textbf{0.104 $\pm$0.038} & 0.146 $\pm$0.030 & 0.105 $\pm$0.031\\
    \midrule
    \multirow{3}{*}{Cowgirl}    & FVD $\downarrow$ & 186.93 & \textbf{174.64}& 210.37 \\
              & PSNR $\uparrow$ & 19.17 $\pm$3.37 & 17.14 $\pm$1.62 & \textbf{24.48 $\pm$2.74}\\
              & LPIPS $\downarrow$ & 0.119 $\pm$0.038 & 0.132 $\pm$0.033 & \textbf{0.091 $\pm$0.031}\\
    \midrule
    \multirow{3}{*}{Survival}  & FVD $\downarrow$ & 196.19 & 236.26 & \textbf{148.50}\\
              & PSNR $\uparrow$  & 23.52 $\pm$3.03 & 15.73 $\pm$1.65 & \textbf{25.55 $\pm$2.73}\\
              & LPIPS $\downarrow$ & 0.087 $\pm$0.035 & 0.158 $\pm$0.038 & \textbf{0.084 $\pm$0.033}\\
    \midrule
    \multirow{3}{*}{Berserker} & FVD $\downarrow$ & 316.14 & 400.99 & \textbf{290.73}\\
              & PSNR $\uparrow$  & 19.56 $\pm$2.42 & 13.34 $\pm$1.48 & \textbf{22.38 $\pm$2.07}\\
              & LPIPS $\downarrow$ & 0.136 $\pm$0.045 & 0.207 $\pm$0.048 & \textbf{0.132 $\pm$0.041}\\
    \midrule
    \multirow{3}{*}{Ghostlady} & FVD $\downarrow$ & \textbf{163.54} & 642.45 & 167.82 \\
              & PSNR $\uparrow$  & 16.85 $\pm$3.69 & 9.84 $\pm$1.88 & \textbf{24.11 $\pm$2.24}\\
              & LPIPS $\downarrow$ & 0.150 $\pm$0.052 & 0.290 $\pm$0.071 & \textbf{0.116 $\pm$0.034}\\
    \midrule
    \multirow{3}{*}{Average}   & FVD $\downarrow$ & 199.93 & 320.06 & \textbf{194.87}\\
              & PSNR $\uparrow$  & 20.31  & 14.54 & \textbf{23.95}\\
              & LPIPS $\downarrow$ & 0.119 & 0.187 & \textbf{0.106} \\
    \bottomrule
  \end{tabular}%
\end{table}

We present additional visual comparisons in \cref{fig:vis_comp} and detailed per-avatar quantitative results in \cref{tab:supp_comparison}. For \methodName{}, due to the inherent stochasticity of diffusion models, the pose of the generated avatar does not exactly match the ground truth even in single-frame generation within each group, which affects Peak Signal-to-Noise Ratio (PSNR) and Learned Perceptual Image Patch Similarity (LPIPS) scores to some extent. Nevertheless, \methodName{} achieves the best performance on the majority of avatars.

For Animatable Gaussians, metrics are computed directly on training SMPL-X parameters. Its worse PSNR and LPIPS scores can be attributed to reduced textural fidelity, as evidenced in \cref{fig:vis_comp}. The underlying challenge is that, for parametric-model-based methods, both appearance quality and motion fidelity depend heavily on precise SMPL-X tracking~\cite{multi_hmr, pixie, osx}. However, accurate SMPL-X estimation—even from multi-view videos—remains an open problem that often demands careful camera calibration and parameter tuning. At the video level, its FVD is comparable to that of \methodName{}; however, this score reflects training-time performance, whereas \methodName{} is evaluated through autoregressive inference.

For Mixamo, although ground-truth skeleton sequences are applied, the resulting avatar poses still differ from the ground truth due to differences in rigging, which accounts for the low image-level scores. On avatars with relatively simple clothing (\textit{Monk} and \textit{Cowgirl}), the influence of non-rigid components is minor, yielding favorable FVD scores. Nevertheless, the robe of \textit{Monk} is broken and the bullets around \textit{Cowgirl}'s waist are twisted, as shown in \cref{fig:vis_comp}. The bag on \textit{Survival} is also unnaturally deformed, revealing an inherent limitation of skeleton-based systems in handling non-body components.

\begin{figure*}[t]
	\centering
	\includegraphics[width=\textwidth]{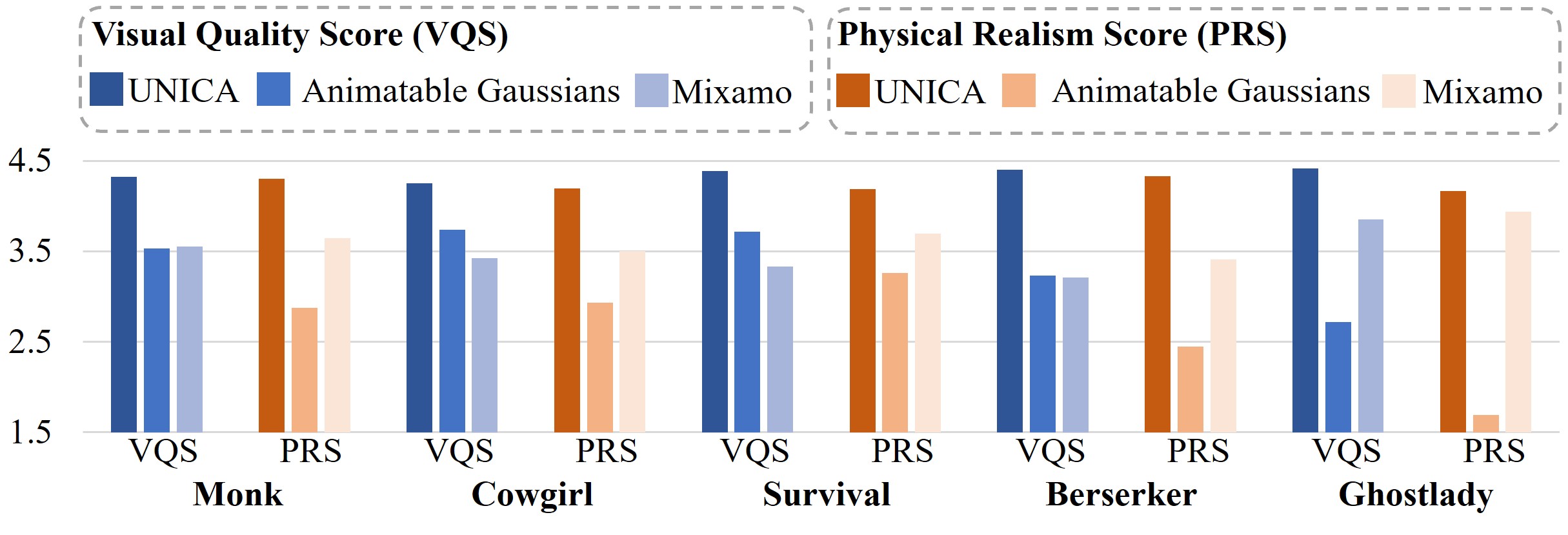}
    \vspace{-0.15in}
    \caption{Per-avatar user study scores on visual quality and physical realism.}
	\label{fig:user_study}
\end{figure*}

\subsection{User Study.} We conducted a user study involving 35 participants, who were asked to rate the Visual Quality Score (VQS) and Physical Realism Score (PRS) of different methods on a scale of 1 to 5. In the study, we presented side-by-side videos of running avatars generated by different methods, with the ground truth provided as a reference. Detailed scores for all five avatars are shown in \cref{fig:user_study}. Our \methodName{} is consistently preferred in terms of both visual quality and physical realism across all avatars. Moreover, \methodName{} scores above 4.0 on both metrics for all five avatars, demonstrating its robustness in capturing the appearance details and physical dynamics of diverse avatars.

\subsection{Runtime Analysis}

\begin{table}[t]
\centering
\vspace{-0.05in}
\caption{Per-frame inference time breakdown and average number of frames required to complete turning events during interactive control.}
\label{tab:runtime}
\setlength{\tabcolsep}{8pt}
\begin{tabular}{cccc|cc}
\toprule
\multicolumn{4}{c|}{Inference Time (s)} & \multicolumn{2}{c}{Frames of Turning} \\
Geometry & Appearance &  PCA & Others & \hspace{0.3em} $180^\circ$ & \hspace{0.3em} $90^\circ$ \\
\midrule
0.414 & 0.250 & $7.23\times10^{-4}$ & $1.85\times 10^{-2}$ & \hspace{0.3em}13.77 & \hspace{0.3em}9.84 \\
\bottomrule
\end{tabular}
\vspace{-0.1in}
\end{table}

\subsubsection{Inference Time.} We report the detailed per-frame inference time of \methodName{} measured on a single NVIDIA 3090 GPU in \cref{tab:runtime}. The geometry inference time comprises 10-step DDIM sampling followed by VAE decoding. Since UNICA operates on four-frame, low-resolution position maps, its computational overhead remains manageable. The appearance inference time corresponds to a single forward pass of PTv3. Despite using a large number of principal components, the computational cost of PCA reconstruction accounts for a negligible fraction of the total runtime. The remaining operations, including progressive 4D inference and Procrustes analysis, are also efficient.

\subsubsection{Responsiveness.} We further evaluate the responsiveness of \methodName{} to action controls by measuring the average number of frames required for the avatar to complete $180^\circ$ and $90^\circ$ turns. To this end, we analyze the trajectories of generated sequences. The analysis is based on the observation that the movement direction between consecutive frames remains consistent when the avatar runs in a fixed direction but transitions gradually during a turning event. We define the number of turning frames as the number of frames required for this directional transition to complete.

In ground-truth motion sequences, $180^\circ$ and $90^\circ$ turns require 12 and 9 frames, respectively. This differs from the 25-frame setting used in the second-stage training of the denoising UNet, as during a turning event, the avatar's movement direction completes changing prior to the stabilization of its body pose. During interactive control, the number of turning frames is expected to be slightly larger, as the avatar cannot initiate a turn while both feet are off the ground. As reported in \cref{tab:runtime}, the measured turning frames remain close to the ground-truth values, confirming the responsiveness of \methodName{}.

\subsection{More Results}
\begin{figure*}[t]
	\centering
	\includegraphics[width=\textwidth]{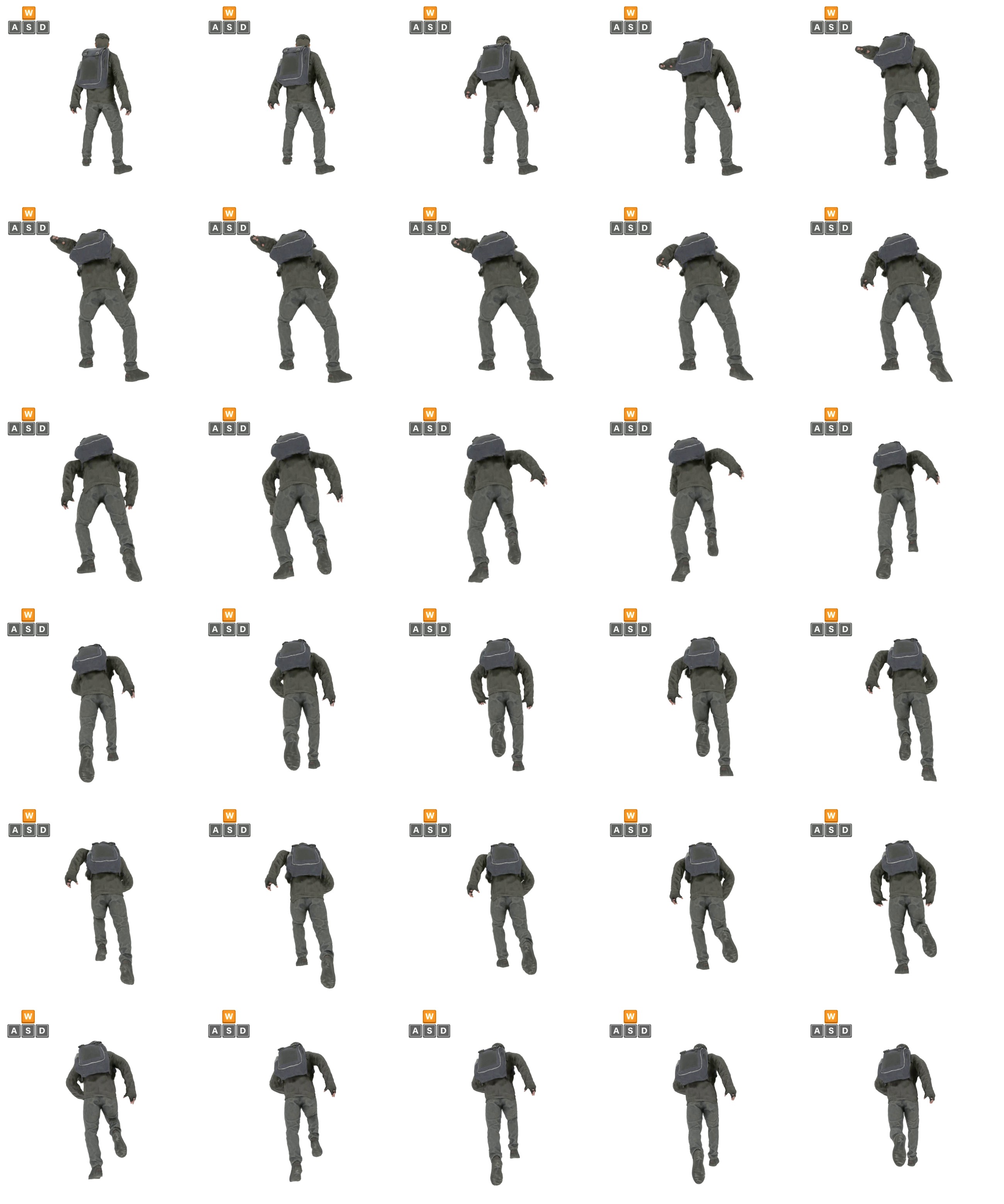}
    \caption{Paired key-rendering results of avatar sprint start. Starting from a standing pose, the avatar starts to run forward under the key \key{W}. The sequence is arranged from left to right, top to bottom.}
	\label{fig:start}
\end{figure*}

\begin{figure*}[t]
	\centering
	\includegraphics[width=\textwidth]{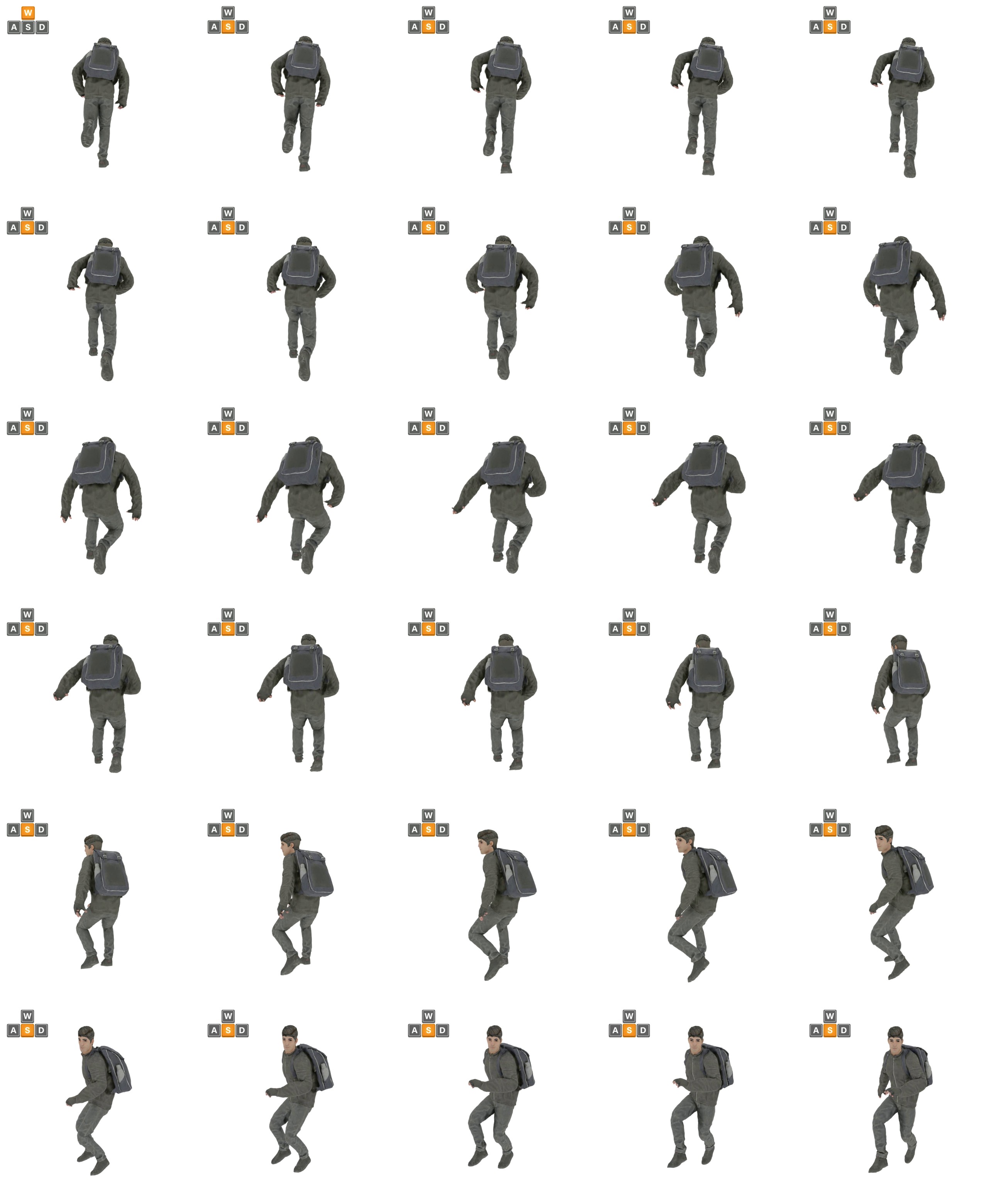}
    \caption{Paired key-rendering results of avatar $180^\circ$ turning, where the key changes from \key{W} to \key{S}. The sequence is arranged from left to right, top to bottom.}
	\label{fig:180-turn}
\end{figure*}

\begin{figure*}[t]
	\centering
	\includegraphics[width=\textwidth]{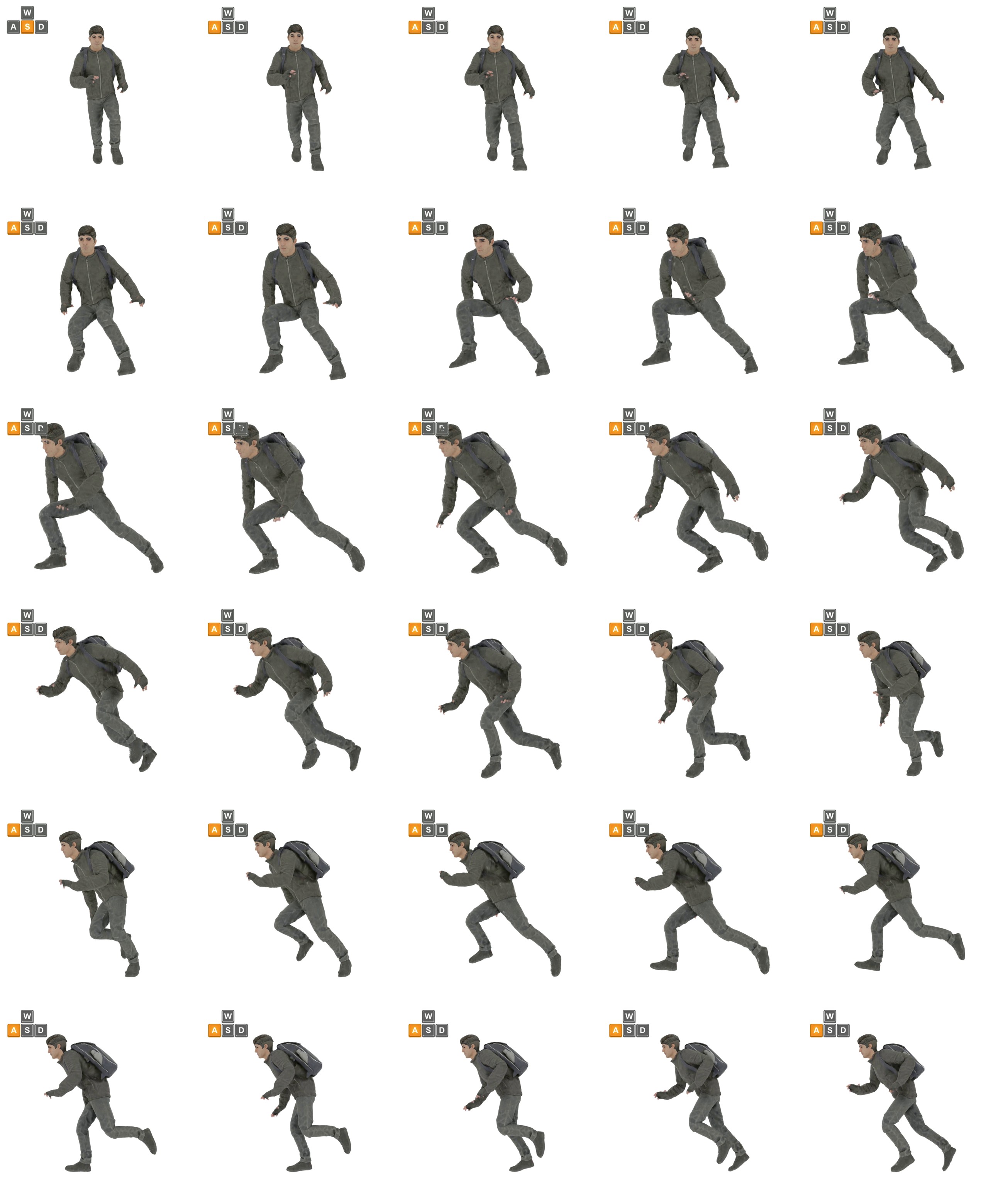}
    \caption{Paired key-rendering results of avatar $90^\circ$ turning, where the key changes from \key{S} to \key{A}. The sequence is arranged from left to right, top to bottom.}
	\label{fig:90-turn}
\end{figure*}

\begin{figure*}[t]
	\centering
	\includegraphics[width=\textwidth]{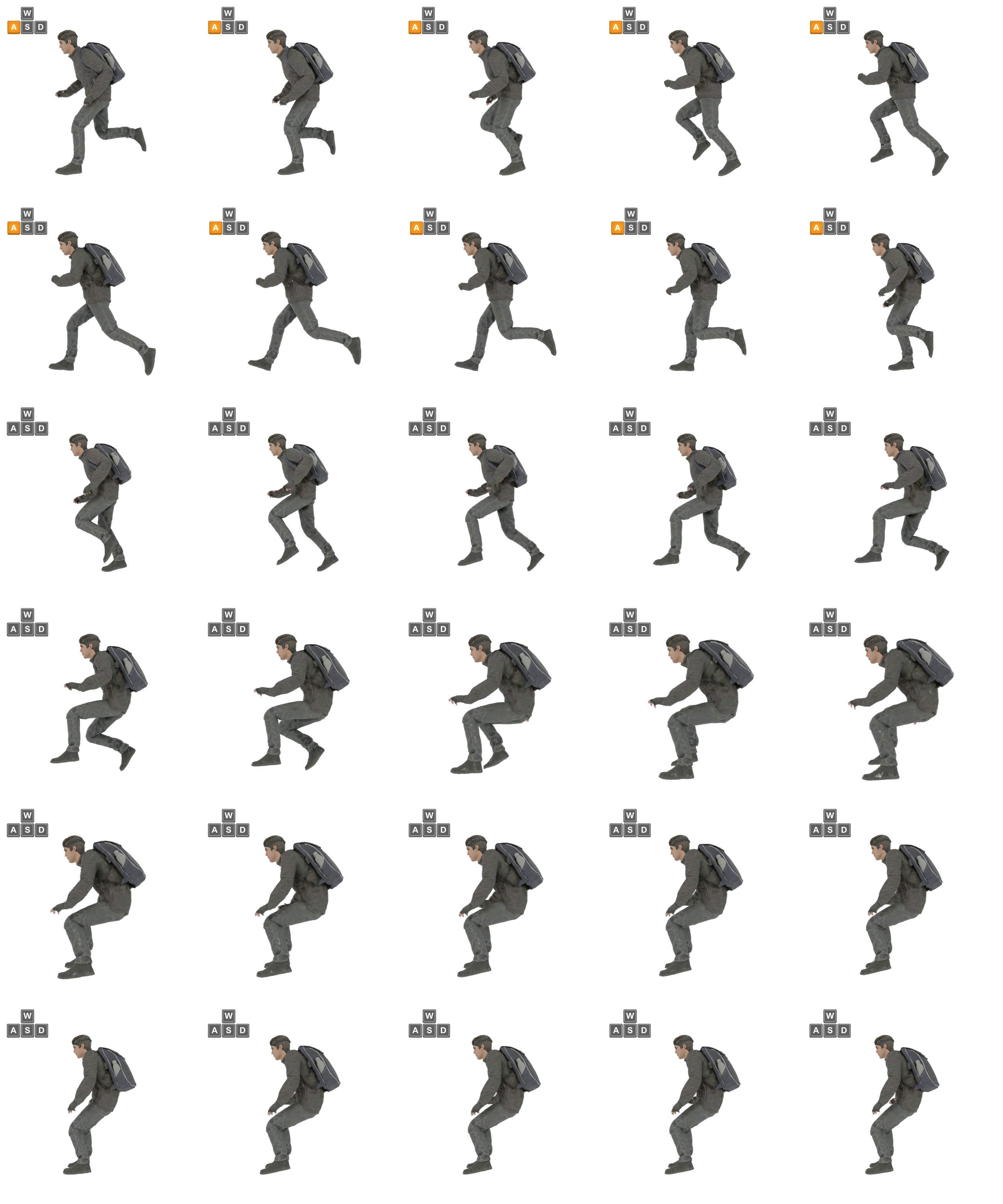}
    \caption{Paired key-rendering results of avatar stopping, where the key changes from \key{A} to \key{$\emptyset$}. The sequence is arranged from left to right, top to bottom.}
	\label{fig:stop}
\end{figure*}

We provide consecutive paired key-rendering results in \cref{fig:start}, \cref{fig:180-turn}, \cref{fig:90-turn}, and \cref{fig:stop} to show how \methodName{} controls the avatar frame-by-frame. Additional sequences can be found in the supplementary video. 

\section{Failure Cases}

\begin{figure*}[t]
	\centering
	\includegraphics[width=0.8\textwidth]{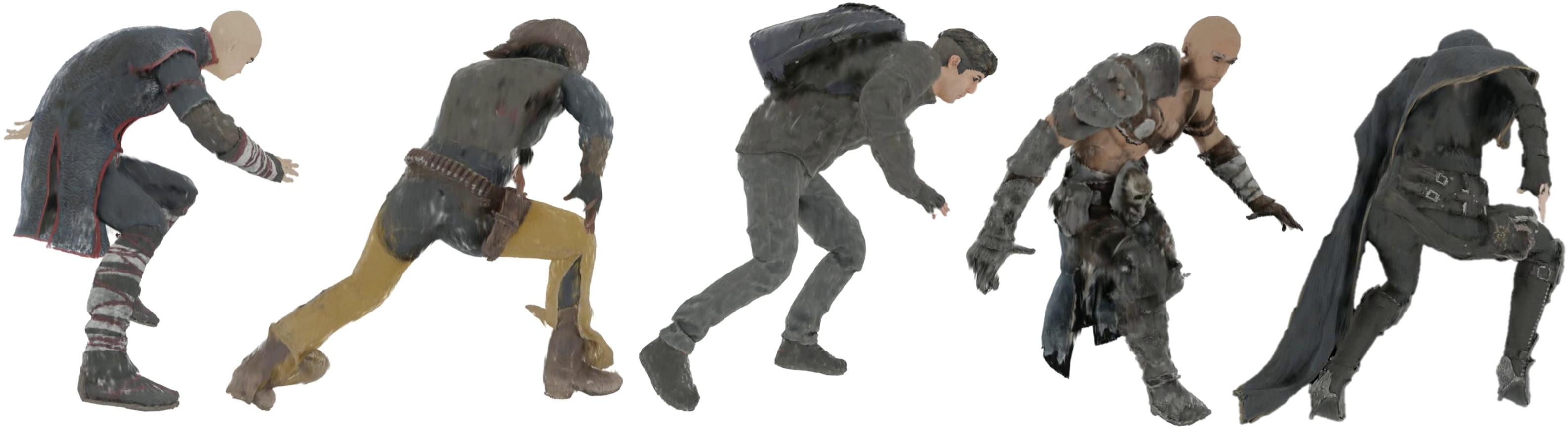}
    \caption{3DGS artifacts observed in turning poses.}
    \vspace{-0.15in}
	\label{fig:limitation}
\end{figure*}

As mentioned in the main paper, the 3DGS renderings of avatars in certain turning poses exhibit artifacts. We provide visualizations of the most prominent cases in \cref{fig:limitation}. Specifically, the 3D Gaussian ellipsoids become dislocated and stretched, resulting in visible gaps between clusters of Gaussians. This problem originates from the appearance mapping stage. We hypothesize that when the avatar leans forward during a sprint start or a turn, the ideal orientations of the Gaussian ellipsoids deviate significantly from those in upright standing or running postures. This discrepancy makes it difficult for the PTv3 model to accurately refine the Gaussian attributes, leading to the reported artifacts. Explicitly rotating the Gaussian ellipsoids according to point normals estimated from neighboring points may help mitigate this issue.

\end{document}